\documentclass[10pt,journal,comsoc]{IEEEtran}
\ifCLASSOPTIONcompsoc
  \usepackage[nocompress]{cite}
\else
  \usepackage{cite}
\fi
\ifCLASSINFOpdf
  \usepackage[pdftex]{graphicx}
\else
\fi
\usepackage{amsmath}
\usepackage{amssymb,amsfonts}

\usepackage{algorithm}
\usepackage{algorithmic}
\usepackage{color}

\begin{document}
\title{GraphTune: A Learning-based Graph Generative Model with Tunable Structural Features}

\author{
    Kohei~Watabe,~\IEEEmembership{Member,~IEEE},
    Shohei~Nakazawa,
		Yoshiki~Sato,
		Sho~Tsugawa,~\IEEEmembership{Member,~IEEE,} 
		and~Kenji~Nakagawa% <-this % stops a space
\IEEEcompsocitemizethanks{\IEEEcompsocthanksitem K.~Watabe, Y.~Sato, and K.~Nakagawa are with the Graduate School of Engineering Nagaoka University of Technology, Nagaoka, Niigata, Japan (e-mail: k\_watabe@vos.nagaokaut.ac.jp; s171039@stn.nagaokaut.ac.jp; nakagawa@nagaokaut.ac.jp).
\IEEEcompsocthanksitem S.~Nakazawa was with the Graduate School of Engineering Nagaoka University of Technology, Nagaoka, Niigata, Japan.
% E-mail: see http://www.michaelshell.org/contact.html
\IEEEcompsocthanksitem S.~Tsugawa is with the Faculty of Engineering, Information and Systems University of Tsukuba, Tsukuba, Ibaraki, Japan (e-mail: s-tugawa@cs.tsukuba.ac.jp).
}% <-this % stops an unwanted space
\thanks{This work was partly supported by JSPS KAKENHI Grant Number JP20H04172.}
\thanks{An earlier and short version of this paper was presented at the 41st IEEE International Conference on Distributed Computing Systems (ICDCS 2021) Poster Track~\cite{Nakazawa2021}. }
% \thanks{Manuscript received April 19, 2005; revised August 26, 2015.}
}

\IEEEtitleabstractindextext{%
\begin{abstract}
% Graph structures are ubiquitous in the real world, 
Generative models for graphs have been actively studied for decades, and they have a wide range of applications. Recently, learning-based graph generation that reproduces real-world graphs has been attracting the attention of many researchers. 
Although several generative models that utilize modern machine learning technologies have been proposed, conditional generation of general graphs has been less explored in the field.
In this paper, we propose a generative model that allows us to tune the value of a global-level structural feature as a condition.
Our model, called GraphTune, makes it possible to tune the value of any structural feature of generated graphs using Long Short Term Memory (LSTM) and a Conditional Variational AutoEncoder (CVAE).
We performed comparative evaluations of GraphTune and conventional models on a real graph dataset. 
The evaluations show that GraphTune makes it possible to more clearly tune the value of a global-level structural feature better than conventional models. 
\end{abstract}

}

\maketitle

\IEEEdisplaynontitleabstractindextext

% \fi
\IEEEpeerreviewmaketitle

\section{Introduction}\label{sec:Introduction}

Generative models for graphs have a wide range of applications, including communication networks, social networks, transportation systems, databases, cheminformatics, and epidemics. 
Repeated simulation on graphs is a basic approach to discovering information in the above fields of study. 
However, researchers and practitioners do not always have access to enough real graph data. 
Generative models of graphs can supplement a graph dataset that does not include a sufficient number of real graphs.
Moreover, generating graphs that are not included in a dataset or future graphs can be used to discover novel synthesizable molecules~\cite{Cao2018,Li2018a,Jonas2019} or to predict the growth of a network~\cite{Bojchevski2018}. 

% 生成が重要という話を書く．生成の使いみち，シミュレーションでの利用など．予測などにも発展する．
% ノードの分類や異常検知も追加する？

% The generative models for graphs can be categorized into two types: statistical models and machine-learning-based models.
Classically, stochastic models that generate graphs with a pre-defined probability of edges and nodes have been studied, and they focus on only a single-aspect feature of graphs.
Various models~\cite{Bonifati2021} have been proposed in the literature, including the Erdős-Rényi (ER) model~\cite{Erdos1959}, Watts-Strogatz (WS) model~\cite{Watts1998}, and Barabási-Albert (BA) model~\cite{Albert2002}. 
These stochastic models accurately reproduce a specific target structural feature (e.g., randomness~\cite{Erdos1959}, small worldness~\cite{Watts1998}, scale-free features~\cite{Albert2002}, and clustered nodes~\cite{Holland1983}). 
In other words, they cannot be adapted to the real data of graphs with numerous features, and cannot guarantee that generated graphs completely reproduce all features of the graphs excluding the target feature. 

% BAなど統計的なモデルの例を説明

Generative models for graphs using machine learning technology learn features from graph data and try to reproduce those features according to the data in every single aspect~\cite{You2018,Li2018,Bojchevski2018,Simonovsky2018,Assouel2018,Yang2019,Lim2020,Goyal2020,Jin2020,Guo2020,Faez2021,Bonifati2021}. 
For the last several years, learning-based graph generation has been attracting the attention of many researchers, and several approaches have been tried in recent studies. 
Although a lot of models have been proposed to generate small graphs with the aim of designing molecules~\cite{Li2018a,Assouel2018,Jonas2019,Lim2020,Jin2020}, some recent studies~\cite{Bojchevski2018,Goyal2020} enable the generation of relatively large graphs that also include citation graphs and social networks. 
In particular, the sequence data-based approach, which converts a graph into sequential data and learns the sequential data by recurrent neural networks, has been successful in the field~\cite{You2018,Li2018,Bojchevski2018,Goyal2020}.
These studies reproduce various features that reflect the global structures of graphs, including average shortest path length, clustering coefficient, and the power-law exponent of the degree distribution.

% 機械学習ベースのモデルで何ができるか説明する．多角的に似せることができるということを書く．多角的に似せた生成は統計モデルではできないことを書く．
% 機械学習的モデルは実データの多面的な特徴を学習して，実データライクなデータを生成できていると書く．

Although the existing generative models for graphs using machine learning technology can generate similar graphs to real-world graphs, most of them cannot generate graphs that have user-specified structural features. 
Although demand for conditional generation of graphs with specific features is common, it has been less explored~\cite{Guo2020,Faez2021}. 

% シミュレーションなどで，特定の特性を調整したいという要求はあるが，多面的に実データに似たものを生成しつつ，特定の特徴量を調整できるモデルはあまりないと書く．既存の機械学習は，基本同じものを生成することを目指していて，調整可能なモデルは殆どないという段落を書く．

Despite the fact that some works aim for conditional generation of graphs with a specific feature, their applicability and performance are not sufficient to generate general graphs with a specific value of a structural feature. % (see Fig.~\ref{fig:position}).
Several models~\cite{Assouel2018,Lim2020,Jin2020} that enable conditional generation utilize domain-specific knowledge of molecule chemistry and are not suitable for graphs of other domains. 
DeepGMG~\cite{Li2018}, one of the pioneering studies of conditional graph generation, does not assume domain-specific knowledge explicitly, and can generate graphs according to a specific condition. 
However, the work just evaluates generation conditioned by the number of atoms (nodes), bonds (edges), or aromatic rings (hexagons) in a molecule. 
DeepGMG does not have an ability to tune global-level structural features (e.g., average shortest path length, clustering coefficient, and the power-law exponent of the degree distribution) which are difficult to tune by adding or removing a local structure of a graph such as a node, edge, or hexagon. 
Although attempts have been made to train DeepGMG by graphs generated by the BA model, they have succeeded in unconditional generation of only very small graphs with 15 nodes~\cite{Li2018}. 
CondGen~\cite{Yang2019}, whose applicable domain is not limited to molecule chemistry, has achieved conditional generation of general graphs including citation graphs. 
CondGen can reproduce global-level structural features (average shortest path length and Gini index are evaluated in the paper~\cite{Yang2019}). 
CondGen succeeds in improving generation by using datasets grouped by label by inputting labels as a condition. 
Unfortunately, it does not provide a model to continuously tune a feature since it requires training datasets grouped by labels.
It does not have sufficient performance to tune the features flexibly based on conditions given as a continuous value of a global-level feature (we discuss the performance of CondGen in Section~\ref{sec:Experiments}).

% ラベルをつけて大域的情報を反映した特徴量に対応できないし，連続的な調整もできない．少しはあるがglobal-levelの構造的特徴は調整できないと書く．DEFactr(Assouel2018), Lim2020, HierVAE(Jin2020) は化学分野．MPGVAEは化学か？% ToDo: GraphVAEを書くべきか．2018なので，CondGenを最新とするか．

% \begin{figure}[tb]
% 	\begin{center}
% 		\includegraphics[width=85mm]{pic/position.pdf}
% 	\end{center}
% 	\caption{position of proposed method}
% 	\label{fig:position}
% 	\vspace{-5mm}
% \end{figure}
% % ToDo: 図を差し替えて文章中で引用

In this paper, we propose GraphTune, a graph generative model that makes it possible to tune the value of any structural feature of a generated graph using Long Short Term Memory (LSTM)~\cite{Hochreiter1997} and a Conditional AutoEncoder (CVAE)~\cite{Kingma2014}.
GraphTune adopts a sequence data-based approach for learning graph structures, and graphs are converted to sequence data by using Depth-First Search (DFS) code that achieves success in GraphGen~\cite{Goyal2020}. 
Unlike GraphGen, GraphTune is a CVAE-based model and the CVAE in the model is composed of an LSTM-based encoder and decoder. 
GraphTune uses CVAE to generate a graph with some specific feature, including global-level structural features, and the feature can be continuously tuned.
Meanwhile, features other than the specified feature are accurately reproduced in every single aspect according to the dataset that is learned. 

% 条件付きベクトルで任意の特徴を与え，それ以外の特徴は実データに似せることができると書く．

In summary, the main contributions of this paper are as follows:
\begin{itemize}
  \item We propose a novel learning-based graph generative model called GraphTune with tunable structural features. 
  In GraphTune, flexible generation of graphs with any feature including global-level structural features (e.g., average shortest path length and clustering coefficient) can be achieved by giving the value of a feature as a condition vector to the CVAE-based architecture. 
  \item We achieve elaborate reproduction of a graph dataset in every single aspect by adopting a sequence data-based approach for learning. 
  GraphTune tunes the value of a specific feature to a specified value while keeping values of other features within the range of values that exist in the dataset.
  % Features of graphs generated by GraphTune are corresponded fairly closely to features of graphs in a learned dataset, except the specified feature. 
  % In other words, with GraphTune, we can tune a feature somewhat independently.
  \item We perform empirical evaluations of GraphTune on a real graph dataset. 
  The evaluation results establish that the tunability and reproducibility of graphs in GraphTune outperforms those in conventional conditional and unconditional graph generative models. 
  The code and dataset used for empirical evaluations are available on GitHub\footnote{https://github.com/wkouw1082/GraphTune}.
\end{itemize}

The rest of this paper is structured as follows. 
In Section~\ref{sec:RelatedWorks}, we summarize related works in the field of graph generative models. 
Section~\ref{sec:ProblemFormulation} formulates the generation problem of a graph with a specified feature. 
In Section~\ref{sec:DFSCode}, we introduce the DFS code that we adopt as a method for converting graphs into sequence data in our model. 
We explain the model architecture and the training and generation algorithms of GraphTune in Section~\ref{sec:GraphTune}. 
Section~\ref{sec:Experiments} shows the empirical evaluations of GraphTune and conventional models. 
Section~\ref{sec:Limitations} discusses the limitations of the paper and future research directions. 
Finally, Section~\ref{sec:Conclusion} concludes the paper.

\section{Related Works}\label{sec:RelatedWorks}

Graph generation has a long history and the literature is rich with the results of many researchers. 
One of the most rudimentary models, Erdős-Rényi model, generates simple random graphs and was proposed in 1959. 
Around 2000, two models~\cite{Watts1998,Albert2002} that reproduce structural features of graphs called small-world networks and power-law degree distribution attracted the attention of researchers.
Since these studies were reported, various statistical graph generation methods inspired by them have been proposed~\cite{Vazquez2003a,Chakrabarti2004,Kolda2014,Bonifati2021}. 
A lot of the structural features of graphs have also been quantified in this research. 
What can be said in common with these traditional statistical generation models is that a model focuses on one (or a few) of the many features and aims to reproduce the features (e.g., small worldness, power-law degree distribution, and local clustering). 
These models cannot be adapted to real graph datasets that have numerous features, and cannot guarantee that the generated graphs completely reproduce every single aspect according to the real data. 

% その後，BAモデルに代表されるスケールフリーなどの特性に着目した研究が数多く提案されていることを書く．NNモデルも書く．
% それらのストリームの中で，グラフを特徴づける様々な統計量が上げられてきた．
% これらの統計的生成に共通して言えることは，数ある特徴量のいくつかに注目し，その注目した特徴を再現することに注力しているということです．
% 多面的に模倣するという考えはない．

A recent trend in the field of graph generation is learning-based models that reproduce real-world graphs. 
Learning-based models have evolved rapidly over the last few years, and have attracted the attention of researchers. 
Learning-based models have been proposed for a wide range of domains ranging from discovering new molecular structures to modeling social networks, and most recently, survey papers have been published~\cite{Guo2020,Faez2021,Bonifati2021}.
Various learning-based models have been proposed, including adjacency-based and edge-list-based approaches.
The sequence data-based approach that converts a graph into sequential data has been particuarly successful in this field~\cite{Li2018,You2018,Bojchevski2018,Goyal2020}.

Although several learning-based models for graphs have been proposed, conditional generation of graphs is less explored~\cite{Guo2020,Faez2021}.
The few models that achieve conditional generation of graphs include domain-specific models in the field of molecule chemistry~\cite{Li2018a,Jonas2019}, evolutionary developmental biology~\cite{Liu2016}, and natural language processing~\cite{Wang2018a,Chen2018}. 
Unfortunately, these models cannot easily be applied to general graphs due to the utilization of domain-specific knowledge.
DeepGMG~\cite{Li2018} and CondGen~\cite{Yang2019} have been proposed as models applicable to general graphs.
DeepGMG adds a condition into the latent vector during the decoding process of the graph to tune the local structure in the graph such as the number of nodes, edges, or hexagons.
CondGen allows tuning of global-level structural features (including average shortest path length and Gini index at least) with graph variational generative adversarial nets.
TSGG-GAN also provides conditional generation of graphs, but it focuses on time series conditioned generation and the challenges are different from ours.
In TSGG-GAN, multivariate time series data are input as node expression values to the model, and graphs conditioned by the time series are generated.

% 数少ないモデルの大部分は化学分野のdomei-specific．

% 変換系は区別して，次の段落で書く．

% NetGAN，MolGAN，GraphGen，"Learning Deep Generative Models of Graphs"，"Constrained Graph Variational Autoencoders for Molecule Design"，"Misc-GAN: A Multi-scale Generative Model for Graphs"あたり

\section{Problem Formulation}\label{sec:ProblemFormulation}

The graphs treated in this paper are undirected connected graphs without self-loop. 
As a notational convention, a graph is represented by $G=(V, E)$, where $V$ and $E$ denote a subset of nodes~$V\subseteq\{v_1, v_2, \dots , v_n\}$ and a subset of edges~$E\subseteq\{(x, y)\,|\,x, y\in V\}$, respectively.
We let $\Omega=\{G_1, G_2, \dots \}$ denote the universal set of graphs $G=(V, E)$. 
% 基本的な記号定義について書く．ラベルなしなので，ラベルありは今後の課題．次数をラベルにしているという話はここでは書かない．グラフの全集合をOmegaにする．非連結なグラフの問題は分ければ同じ問題．

\begin{figure}[tb]
  \begin{center}
    \includegraphics[width=0.9\linewidth]{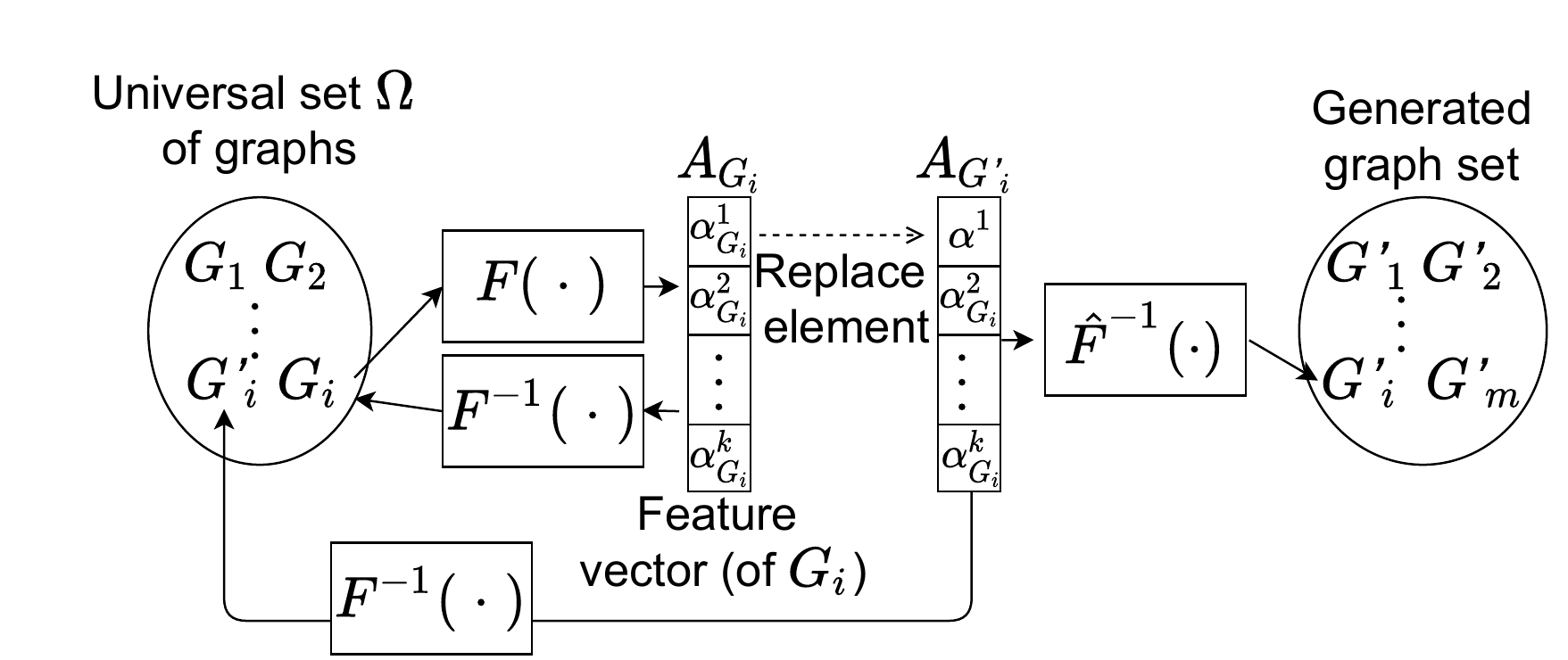}
  \end{center}
  \caption{Problem formulation. 
  A graph $G_i\in \Omega$ is mapped to a feature vector $A_{G_i}\in \boldsymbol{A}$ by a mapping $F(G_i)=A_{G_i}=[\alpha_{G_i}^1 \alpha_{G_i}^2 \cdots]^T$.
  Elements of a feature vector $A_{G_i}$ for graph $G_i$ represent values of all sorts of features that are calculated from $G_i$.
  This paper tackles the inference problem of finding $\hat{F}^{-1}(\cdot)$ that approximates $F^{-1}(\cdot)$, using a subset $\mathcal{G}$ of the universal set $\Omega$ of graphs.
  $\hat{F}^{-1}(\cdot)$ can generate $G_i'$ by inputting the feature vector $A_{G_i'}$ in which any element of the vector $A_{G_i}$ is replaced by an arbitrary value.}
  \label{fig:problem_formulation}
\end{figure}

We consider a mapping $F: \Omega\to \boldsymbol{A}$, and a graph $G_i\in \Omega$ that is mapped to a feature vector $A_{G_i}\in \boldsymbol{A}$ by $F(G_i)=A_{G_i}=[\alpha_{G_i}^1 \alpha_{G_i}^2 \cdots]^T$ as shown in Fig.~\ref{fig:problem_formulation}.
The $j$th element $\alpha_{G_i}^j$ of the vector $A_{G_i}$ expresses a feature of graph $G_i$.
Every feature is represented by a real number $\alpha_{G_i}^k\in \mathbb{R}$. 
% We assume that a feature vector $A_{G_i}$ for graph $G_i$ contains values for all the features that are calculated from $G_i$ (i.e., $A_{G}=A_{G'}\Rightarrow G=G'$). 
For example, elements of $A_{G_i}$ represent values of features such as the number of nodes and edges, average shortest path length, average degree, edge density, modularity, clustering coefficient, power-law exponent of the degree distribution, and largest component size.
% Aはグラフのあらゆる特徴を表すベクトルであり，G_iと一意に対応するということを書く．Fig.1を参照する．

In this paper, we formulate the problem of generating a graph with specified features as inferencing the inverse image from feature vectors of graphs to graphs (see Fig.\ref{fig:problem_formulation}).
We define the inverse image $ F^{-1} (\cdot) $ of the mapping $F(\cdot)$, and graph $G_i$ can be obtained by calculating $F^{-1} (A_{G_i}) = \{G_i, \dots\} $ with the inverse image.
We tackle the inferencing problem of finding $\hat{F}^{-1}(\cdot)$ that approximates $F^{-1}(\cdot)$ by using a subset $\mathcal{G}$ of $\Omega$.
By solving this problem, it is possible to generate $G_i'$ with the feature vector $A_{G_i'}$ in which any element of the vector $A_{G_i}$ is replaced by an arbitrary value.
Since we cannot use the universal set $\Omega$ of graphs, the inference needs to be achieved by a subset $\mathcal{G}$ of $\Omega$ that is contained in an accessible dataset.
% Fがわかれば生成できるということを書く．
% ■■■ inverse mapping ではなく inverse image として書き直す ■■■ 済

%グラフ$G_i$は, 写像$F(\cdot)$によって, $F(G_i)=A_{G_i}=[\alpha_{G_i}^1 \alpha_{G_i}^2 \cdots \alpha_{G_i}^k]^T$のベクトルに射影される.
%ベクトル$A_{G_i}\in \mathbb{R}^k$は, 入力されたグラフ$G_i$の特徴を表現したベクトルになっており, ベクトルの各次元の値がそれぞれ, 別個の構造的特徴量(グラフのノード数, グラフのエッジ数, 次数分布のべき指数, クラスタ係数等)の値になっている. 
%ここで, 逆写像$F^{-1}(\cdot)$を定義し, 逆写像により, $F^{-1}(A_{G_i})=G_i$であるとする. 

%本稿では, 図\ref{fig:problem_formulation}のように, 存在しうる全てのグラフ集合の部分集合である$\mathcal{S}=\{G_1, \dots, G_m\}\subset \bm{\Omega}$より, $F(\cdot), F^{-1}(\cdot)$を近似した$\hat{F}(\cdot), \hat{F}^{-1}(\cdot)$を, 機械学習を用いて学習する問題を取り扱う.  
%この問題を解くことにより, ベクトル$A_{G_i}$の任意の次元を任意の値に変更したベクトル$A_{G_i'}$に従う$G_i'$を構築することが可能である. 
%例えば, ベクトル$A_{G_i}$における$1$次元目の要素を$\alpha^1$に変更した$A_{G_i'}=[\alpha^1 \alpha_{G_i}^2 \cdots \alpha_{G_i}^k]^T$より, $\hat{F}^{-1}(A_{G_i'})=G_i'$を達成することが可能である. 

\section{DFS Code}\label{sec:DFSCode}

GraphGen~\cite{Goyal2020} is a successful model for unconditional generation in learning-based graph generation that utilizes the DFS code. 
The key idea is to use the DFS code to convert graphs into sequence data. 
% In the preprocess for training in GraphGen, graphs in a dataset are converted to sequence data. 
The converted sequence data are learned by LSTM in a training process. 
The compact expression of a graph by sequence data with DFS code allows GraphGen to accurately generate graphs that are similar to graphs in a dataset.
While we propose a novel CVAE-based generative model for graphs different from GraphGen, our model follows the sequence data-based approach by using DFS code in the preprocess for training.
In this section, we summarize the conversion process by DFS code that is common to GraphGen and our model.
%GraphGenが一つのLSTMでシーケンスを学習していることを書いた上で，我々がシーケンシャル学習のアイデアをCVAEでに組み合わせることで条件付き生成ができるようにしたとうことを書く．本章ではDFSコードについて書くと書く．

DFS code converts a graph to a unique sequence of edges that retains the structural features of the graph.
It is well known that a trajectory (i.e., a sequence) of a walk on a graph reflects features of the graph, including the degree distribution~\cite{Spielman2011}.  % 固有値を入れても良い
DFS code converts a graph to a compact sequence of length $|E|$ by using the depth-first search preventing revisit of edges. 
% エッジシーケンスに変換することの良さを書く．ランダムウォークが次数分布や固有値を始めグラフの特徴を反映することはよく知られている．

%GraphGenは, グラフの隣接行列により, グラフの構造的特徴を捉えるのではなく, グラフをユニークなシーケンスに変換し, シーケンスの学習を行う. 
%一つ目として, 隣接行列の計算量は$|V|^2$に依存するのに対して, シーケンスへの変換を行う, $|E|$依存にすることが可能である. 
%二つ目として, グラフを規則的に一対一対応のシーケンスへの変換が可能である点である. 
%DFSコードは, グラフ上を深さ優先探索した際のエッジの探索順序のシーケンスにエンコードする手法である. 
%深さ優先探索は, グラフのどのノードから始めても良い. 

\begin{figure}[tb]
  \begin{center}
    \includegraphics[width=0.8\linewidth]{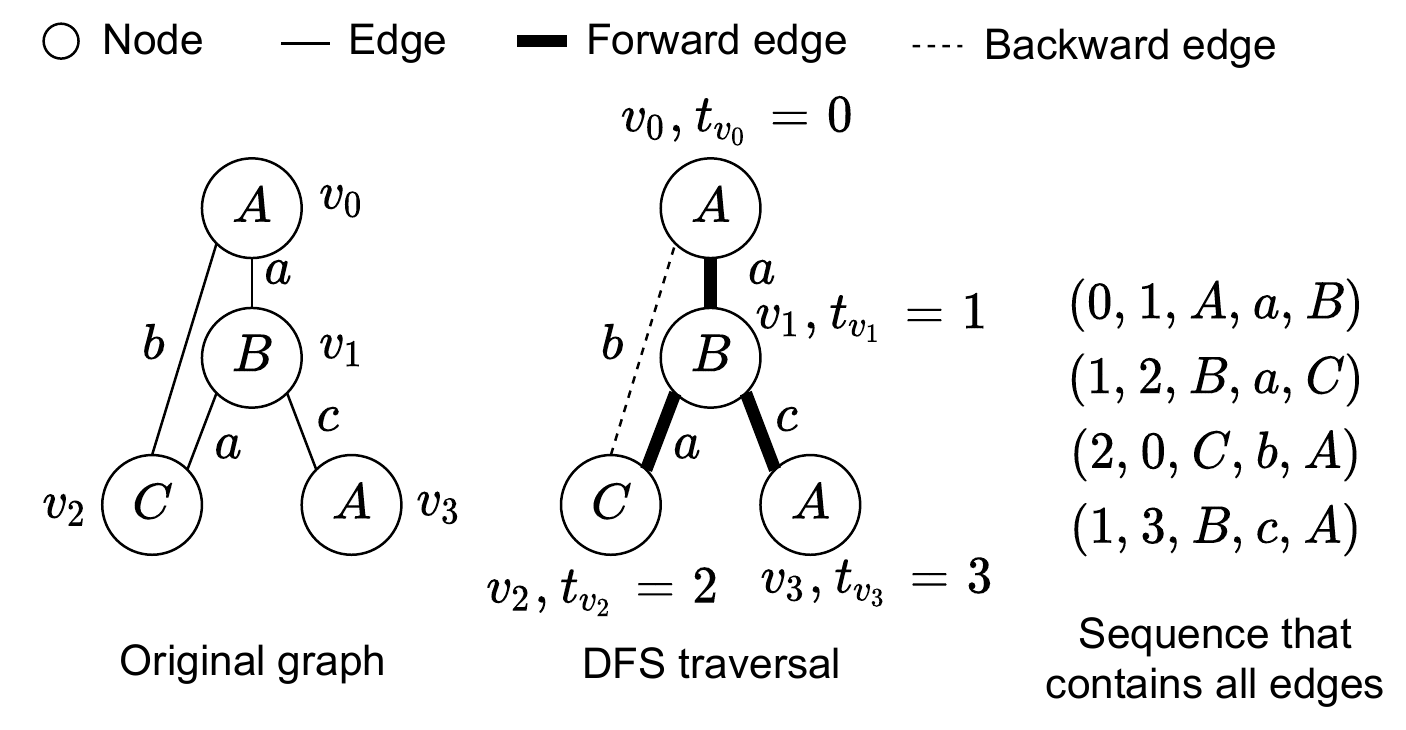}
  \end{center}
  \caption{Sequence converted from an example graph. All nodes $v_i (i=0,1,\dots)$ are assigned timestamps $t_{v_i}$ in order of depth-first search, thereby obtaining a DFS traversal. 
  The edges contained in the DFS traversal are called forward edges, and other edges are called backward edges. 
  A backward edge $(u, u')$ is placed between forward edges $(w, u)$ and $(u, v)$, thereby obtaining a sequence that contains all edges.}
  \label{fig:traversal}
  \vspace{-5mm}
\end{figure}

In the conversion algorithm of DFS code, timestamps are first added to all nodes from 0 by performing the depth-first search.
That is, all nodes $v_i (i=0,1,\dots)$ are discovered and assigned timestamps $t_{v_i}$ in order of depth-first search. 
The traversal of the search is represented as a sequence of edges called a DFS traversal. 
In the example graph shown in Fig.~\ref{fig:traversal}, a DFS traversal is $(v_0, v_1), (v_1, v_2), (v_1, v_3)$, and  timestamps of nodes are assigned as $t_{v_0}=0$, $t_{v_1}=1$, $t_{v_2}=2$, and $t_{v_3}=3$. 
Edges contained in a DFS traversal are called forward edges, and other edges are called backward edges. 
By adding a timestamp to all nodes, edge $e=(u, v)$ can be annotated as a 5-tuple $(t_u, t_v, L(u), L(e), L(v))$, where $t_u$ and $L(\cdot)$ denote the timestamp of node $u$ and the label of the edge or node, respectively. 
Although the graphs treated in this paper are graphs with unlabeled nodes and edges, the original DFS code is designed for labeled graphs. 
A detailed treatment of $L(u)$, $L(e)$ in our model is explained in Section~\ref{ssec:SecuentialData}. 
% Fig traversalを参照．forward edgeやbackward edgeの定義を説明．この研究ではラベルなしグラフを想定しているが，オリジナルのDFSコードではラベルを考慮した設計になっていると書く．

%DFSコードでは，エッジのシーケンスにエンコードするが, それに先立ち，ノードを深さ優先探索で順番付け，0から順番にタイムスタンプを付与する. 
%エッジ$(u, v)$に訪れた際, ノード$u, v$にそれぞれタイムスタンプ$t_u, t_v$を割り振る.  
%図\ref{traversal}(a)のようなグラフが存在した際に, 深さ優先探索の順序が$(v_0, v_1), (v_1, v_2), (v_1, v_3)$であった際のタイムスタンプは図\ref{traversal}(b)のように示される. 
%$t_u < t_v$, すなわち, ノードの順番付けと同じ方向につながるエッジをforward edge, $t_u > t_v$,  すなわちノードの順番付けを遡る方向のエッジをbackward edge と呼ぶ. 
%エッジ$e=(u, v)$はタイムスタンプを定義したことにより, 5-tupleである$(t_u, t_v, L(u), L(e), L(v))$で記述することが可能である. 

Based on the order of node timestamps, DFS code constructs a sequence that contains all edges in a graph. 
Although the forward edges are already constructed as a sequence (i.e., a DFS traversal), the backward edges are not included in the sequence. 
To construct a sequence that contains all edges in a graph, a backward edge $(u, u')$ is placed between forward edges $(w, u)$ and $(u, v)$. 
If there are multiple backward edges $(u, u')$ and $(u, u'')$, the timestamps of $u'$ and $u''$ are compared, and the smaller one is placed in front. 
By performing this procedure, all backward edges are placed between forward edges, and a sequence that contains all edges is obtained. 
In the example shown in Fig.~\ref{fig:traversal}, the obtained sequence is $(0, 1, A, a, B), (1, 2, B, a, C), (2, 0, C, b, A), (1, 3, B, c, A)$. 
% DFS code converts the graph so that the sequence composed of consecutive neighboring edges as much as possible while maintaining the order of timestamps of the nodes.
Although sequences that are constructed by the above procedure are not necessarily unique, the lexicographically smallest sequence based upon lexicographical ordering~\cite{Yan2002} is chosen as the unique one. 
As a result, the graph is represented as unique sequence of 5-tuple $(t_u, t_v, L(u), L(e), L(v))$ by DFS code.
% ユニークにするための工夫を説明しておく．例も書く．

It is not difficult to reconvert a DFS code into a graph if the DFS code exactly represents the graph. 
By constructing the edges and the nodes on either side of these edges in the order listed in the DFS code, we can reconstruct the graph.
In the case of the example shown in Fig~\ref{fig:traversal}, the graph can be reconstructed by the following procedure.
1) Create Node 0, Node 1, and Edge 0-1 with labels $A$, $B$, and $a$ respectively; 
2) Create new Node 2 and Edge 1-2 with labels $C$ and $a$, respectively; 
3) Create new Edge 2-0 with label $b$. 
4) Create new Node 3 and Edge 1-3 with labels $A$ and $c$, respectively. 

\section{GraphTune: A Graph Generative Model with Tunable Structural Features}\label{sec:GraphTune}

We propose \textit{GraphTune} -- a generative model for graphs that is able to tune a specific structural feature using DFS code and CVAE.
GraphTune is composed of CVAE with an LSTM-based encoder and decoder. 
Graphs are converted to sequence data by the DFS code, and the sequence data are input to LSTM. 
This section provides a detailed review of generative approaches of graphs in GraphTune. 
% 簡単なGraphTuneの要約(シーケンスデータを学習することや，エンコーダとデコーダにLSTMを使うなど)と，GraphTuneを本章で紹介すると書く．

%提案モデルでは, GraphGenをConditional Vartiational Auto Encoder (CVAE)で拡張し, グラフの構造的特徴の値を条件としたベクトルに従ったグラフを生成可能なモデルを提案する. 
%本章では, 提案法のモデルの概要を示す. 
% The abstract of proposed model is as shown in section.

\subsection{Sequence Data Converted from Graphs}\label{ssec:SecuentialData}

Like GraphGen, GraphTune learns a sequence dataset $\mathcal{S}=\{F_{\rm DFS}(G_i) | G_i\in \mathcal{G}\}$ that is converted from a graph dataset $\mathcal{G}\subset\Omega$ using a conversion $F_{\rm DFS}(\cdot)$ based on DFS code. 
Although the conversion $F_{\rm DFS}(\cdot)$ is basically the same as the conversion by DFS code described in Section~\ref{sec:DFSCode}, it is modified as follows to adapt it for our problem. 
As mentioned above, although the original DFS code is designed for labeled graphs, we assume unlabeled graphs in Section~\ref{sec:ProblemFormulation}. 
In the sequence dataset that is learned by GraphTune, node labels $L(v)\,(v\in V)$ and edge labels $L(e)\,(e\in E)$ in a 5-tuple are set as the degree of node $v$ and 0, respectively. 
According to the DFS code procedure described in Section~\ref{sec:DFSCode}, a graph is converted to a sequence of 5-tuples.
At the end of the sequences, an End Of Sequence (EOS) token $(\mathrm{EOS}_{t}, \mathrm{EOS}_{t}, \mathrm{EOS}_{L}, 1, \mathrm{EOS}_{L})$ is added, where $\mathrm{EOS}_{t}$ and $\mathrm{EOS}_{L}$ represent $1+\max_{v\in \boldsymbol{V}}t_v$ and $1+\max_{v\in \boldsymbol{V}}L(v)$ for a set $\boldsymbol{V}$ of all nodes in the graph dataset $\mathcal{G}$. 
Sequences with EOS allow us to learn graphs of any size.  
We can obtain a sequence $\boldsymbol{S}_i\in\mathcal{S}$ by further converting the sequence of 5-tuples by component-wise one-hot encoding. 
A sequence $\boldsymbol{S}_i$ with element $\boldsymbol{s}_j\in\{0, 1\}^k$ is input into the model of GraphTune (see below for the model architecture).

% EOSを追加するなどを説明する．ラベルはエッジラベルは全部ゼロ，ノードラベルは次数にしているということを書く．DFSコードはラベル付きデータの学習を可能にする構造にあるが，それは今後の課題と書く．

%提案モデルでは, GraphGenと同様にグラフ$\bm{G}$をDFSコードによって変換したシーケンスデータセット$\mathcal{S}=F_{\rm DFS}(\bm{G})$を学習する. 
%DFSコードのシーケンスの最後の要素は, シーケンスの終了EOSを表すベクトルとなっている. 
%本稿では, グラフデータはノード, エッジ共にラベルを有していないものとし, 前処理で, ノードに次数の逆数をラベリングしている. 

\subsection{Condition Vectors Corresponding to Graphs}\label{ssec:ConditionVectors}

Along with the sequential data fed by the DFS code, we input condition vectors $\mathcal{C}$ expressing structural features of the graph dataset $\mathcal{G}$ to the GraphTune model for learning. 
Elements $c_i$ of vector $\boldsymbol{C}_i=[c_i, c_i, \dots, c_i]^T\in\mathcal{C}$ represent the value of the structural feature of graph $G_i$ in graph dataset $\mathcal{G}$, and $c_i$ corresponds to the first element of a feature vector $A_{G_i}$ in Fig.\ref{fig:problem_formulation}. 
Elements $c_i$ of condition vector $\boldsymbol{C}_i$ are calculated from graph $G_i$ by a statistical process. 
The condition vector specifies the structural features we focus on, and we can choose any structural feature as the elements of a condition vector.
Structural features that can be specified in a condition vector are not limited to features regarding the local structures of graphs such as the number of nodes and edges, but global-level structural features (including average of shortest path length, clustering coefficient and the power-law exponent of the degree distribution) can also be specified.
For example, if we want to tune the model by focusing on the clustering coefficient of the graphs, then we calculate a clustering coefficient $\alpha_\mathrm{cls}(G_i)$ for each graph $G_i\in\mathcal{G}$ and construct a vector $[\alpha_\mathrm{cls}(G_i), \alpha_\mathrm{cls}(G_i), \dots , \alpha_\mathrm{cls}(G_i)]$ for each $G_i\in\mathcal{G}$. 

% ベクトルが$A_{G_i}$の一部であることを書く．注目する統計量を任意に選べること．専攻手法がやっているようなローカルな特徴やノード数やエッジ数のようなシンプルなものではなく，大域的な計算が必要な統計量であることを書く．

%グラフの次数分布のべき指数を計算する関数を$F_{\rm degree}(\cdot)$, グラフのクラスタ係数を計算する関数を$F_{\rm cluster}(\cdot)$とした時, これらの関数を用いて, グラフのデータセットの集合に対応した条件ベクトルの集合${\bm C}=\{c_1, \dots, c_m\}=\{[F_{\rm degree}(G_1)\; F_{\rm cluster}(G_1)]^T, \dots, [F_{\rm degree}(G_m)\; F_{\rm cluster}(G_m)]^T\}$を作成する. 

\subsection{Model Architecture}\label{ssec:ModelArchitecture}

\begin{figure}[tb]
  \begin{center}
    \includegraphics[width=\linewidth]{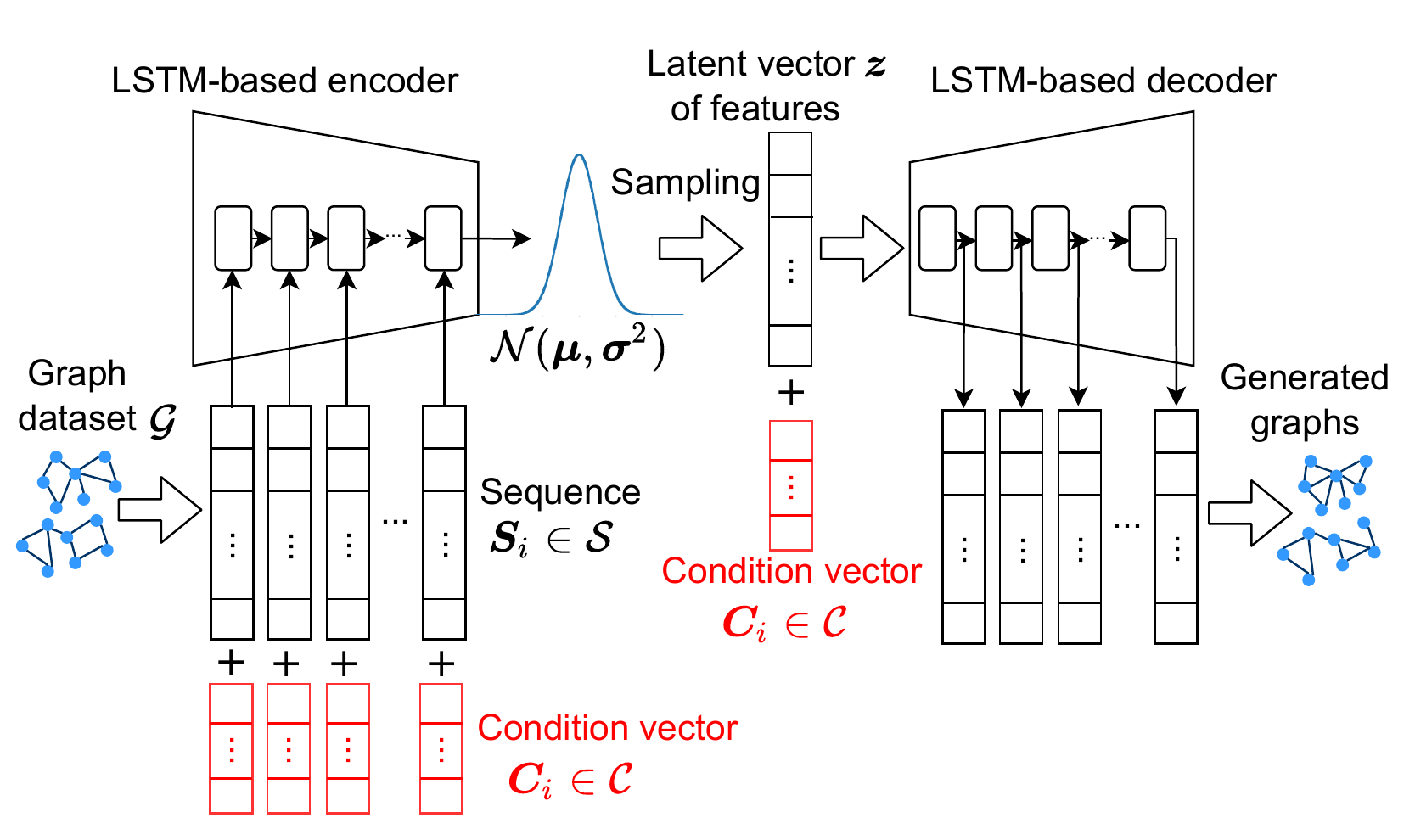}
  \end{center}
  \caption{Proposed model composed of CVAE with a LSTM-based encoder and decoder. 
  A graph in the graph dataset $\mathcal{G}$ is converted to a sequence $S_i\in\mathcal{S}$ by using DFS code. The sequence is processed by an LSTM-based encoder. 
  The decoder generates a sequence of 5-tuples, and the sequence is converted to a generated graph. 
  The condition vector is input to both the encoder and decoder. 
  See the equations in Section~\ref{ssec:ModelArchitecture} for the detailed process of the model. }
  \label{fig:model}
\end{figure}

The proposed model is composed of CVAE with an LSTM-based encoder and decoder (see Fig.~\ref{fig:model}). 
A graph dataset $\mathcal{G}$ is converted to a sequence dataset $\mathcal{S}$ in the manner explained in Section~\ref{sec:DFSCode}. 
By doing this, set of condition vectors $\mathcal{C}$ is calculated from the graph dataset $\mathcal{G}$. 
The proposed model is trained with a sequence dataset $\mathcal{S}$ and a condition vector set $\mathcal{C}$. 
Sequence data ${\boldsymbol{S}_i\in\mathcal{S}}$ and the condition vector $\boldsymbol{C}_i\in\mathcal{C}$ are input into the LSTM-based encoder, and the encoder finds a latent state distribution $\mathcal{Z}$ of latent vectors.
A latent vector $\boldsymbol{z}$ is randomly sampled from the distribution $\mathcal{Z}$, and the latent vector $\boldsymbol{z}$ concatenated with a condition vector $\boldsymbol{C}_i\in\mathcal{C}$ is input into the LSTM-based decoder. 
The decoder tries to reproduce the sequence $\boldsymbol{S}_i$. 
The details of the encoder and the decoder are described below. 
% 全体的な構成とデータの流れを簡単に書く．

%提案モデルは, EncoderとDecoderの二種類のLSTMから成るVAEであり, 図\ref{model_pic}のように表される. 
%前処理を行ったDFSコードによるシーケンスデータセット$\mathcal{S}$と条件ベクトルの集合$\mathcal{C}$を用いて学習を行う.

\subsubsection*{Encoder}
Encoder~$q_{\boldsymbol{\phi}}(\boldsymbol{z}|\boldsymbol{S}_i)$ with parameter $\boldsymbol{\phi}$ learns sequences $\boldsymbol{S}_i$ and maps them to a latent vector $\boldsymbol{z}$ according to features of graphs. 
To treat sequence data, we employ a stacked LSTM as an encoder. 
% A condition vector $\boldsymbol{C}_i$ is fed to the stacked LSTM as an initial hidden state vector $\boldsymbol{h}_0$ through a fully connected layer $f_\mathrm{einit}$.
% ToDo: 定式化では\boldsymbol{C}_iの供給ははないが，条件は\boldsymbol{S}_iに含まれている情報であり，明にエンコーダに供給することで効率的な学習を期待できるということを書く．
To encode a graph into a latent space, the $j$th element $\boldsymbol{s}_j$ of the sequence $\boldsymbol{S}_i$ and a condition vector $\boldsymbol{C}_i$ are vertically concatenated in a single vector and the vector is embedded with a single fully connected layer $f_\mathrm{eemb}$.
The embedded vector is then fed into each LSTM block $f_\mathrm{enc}$.
The initial hidden state vector $\boldsymbol{h}_0$ is initialized as a zero vector $\boldsymbol{0}$. 
The stacked LSTM with the embedding layer $f_\mathrm{eemb}$ processes a sequence $\boldsymbol{S}_i$ of length $k=|\boldsymbol{S}_i|$ by recursively applying the LSTM block $f_\mathrm{enc}$ to hidden state vector $\boldsymbol{h}_j$. 
The output $\boldsymbol{h}_k$ of the last LSTM block is fed to two functions $f_{\boldsymbol{\mu}}$ and $f_{\boldsymbol{\sigma}^2}$ implemented by single fully connected layers.
As usual in VAE, the latent state distribution is enforced as a multivariate Gaussian distribution with dimension $L$. 
A latent vector $\boldsymbol{z}$ is then sampled from the latent state distribution $\mathcal{N}(\boldsymbol{\mu}, \boldsymbol{\sigma}^2)$ where $\boldsymbol{\mu}=f_{\boldsymbol{\mu}}(\boldsymbol{h}_k)$ and $\boldsymbol{\sigma}^2=f_{\boldsymbol{\sigma}^2}(\boldsymbol{h}_k)$. 
Summarizing the above, the process of the encoder part of our model is as follows: 
\begin{align}
  % \boldsymbol{h}_0 &= f_\mathrm{einit}(\boldsymbol{C}_i), \\
  \boldsymbol{h}_0 &= \boldsymbol{0}, \\
  \boldsymbol{h}_{j+1} &= f_{\rm enc}(\boldsymbol{h}_{j}, f_{\rm eemb}([\boldsymbol{s}_{j}^T, \boldsymbol{C}_i^T]^T)) \;\; (j=0, 1, \dots , k-1), \label{eq:enc_cond}\\
  \boldsymbol{\mu} &= f_{\boldsymbol{\mu}}(\boldsymbol{h}_k)\\
  \boldsymbol{\sigma}^2 &= f_{\boldsymbol{\sigma}^2}(\boldsymbol{h}_k)\\
  \boldsymbol{z} &\sim \mathcal{N}(\boldsymbol{\mu}, \boldsymbol{\sigma}^2).
\end{align}

%{\bf{Encoder}} EncoderはLSTMを用いており, DFSコードによるシーケンスを学習し, その特徴に応じた多変量正規分布にマッピングを行い,  その分布から潜在変数$\boldsymbol{z}$をサンプリングする.
%グラフ$G_i$のDFSコードによるシーケンスを$F_{\rm DFS}(G_i)=\boldsymbol{S}_i=\{\boldsymbol{s}_1, \boldsymbol{s}_2, \dots, \boldsymbol{s}_k\}$, 条件ベクトルを$c_i$とする. 
%Encoderを$f_{\rm enc}$, 埋め込み関数を$f_{\rm eemb}$とした時, 次式に示すように, 学習時は, Encoderに, シーケンスの各要素に対して$c_i$をvstackしたベクトルを再帰的に入力する. 
%$k(=|\boldsymbol{S}_i|)$回のイテレーションの後に, LSTMの出力である$\boldsymbol{h}_k$を, 線形層$f_{\boldsymbol{\mu}}, f_{\boldsymbol{\sigma}}$を用いて, 多変量正規分布のパラメータに変換し, reparameterization trickによって, グラフ$G_i$の潜在変数である$\boldsymbol{z}$に変換する. 
%$c_i$と$\boldsymbol{s}_{j-1}$をEncoderに対してともに入力することで, Encoderに対して, 現在の入力$\boldsymbol{s}_{j-1}$の条件$c_i$であることを学習させる.

\subsubsection*{Decoder}
Decoder~$p_{\boldsymbol{\theta}}(\boldsymbol{S}_i|\boldsymbol{z})$ with parameter $\boldsymbol{\theta}$ learns to map a subsequence $\{\boldsymbol{s}_m|m\le j\}$ of a sequence $\boldsymbol{S}_i$, a condition vector $\boldsymbol{C}_i$, and latent vector $\boldsymbol{z}$ to a next element $\boldsymbol{s}_{j+1}$. 
The decoder is also modeled by a stacked LSTM. 
% A sampled latent vector $\boldsymbol{z}$ and the condition vector $\boldsymbol{C}_i$ are concatenated together into a single vector, and the concatenated vector is fed to the stacked LSTM as an initial hidden state vector $\boldsymbol{h}_0$ through a fully connected layer $f_\mathrm{dinit}$.
Like the encoder, the $j$th element $\boldsymbol{s}_j$ of the sequence $\boldsymbol{S}_i$ is embedded into a vector by a single fully connected layer $f_\mathrm{demb}$, and is processed by a stacked LSTM.
Unlike the encoder, however, the embedded vector is concatenated with a sampled latent vector $\boldsymbol{z}$ and a condition vector $\boldsymbol{C}_i$ before inputting into a stacked LSTM.
The concatenated vector is fed into the LSTM block $f_\mathrm{dec}$, and a sequence $\boldsymbol{S}_i$ is processed by recursively applying the LSTM block $f_\mathrm{dec}$. 
The initial hidden state $\boldsymbol{h}_0$ is calculated by replacing $\boldsymbol{s}_j$ and $\boldsymbol{h}_j$ with a Start Of Sequence ($\mathrm{SOS}$) and $f_\mathrm{dinit}(\boldsymbol{C}_i)$. 
$\mathrm{SOS}$ is converted by $f_\mathrm{dsos}$ from a vector into which a latent vector $\boldsymbol{z}$ and a condition vector $\boldsymbol{C}_i$ are concatenated together, where the conversion $f_\mathrm{dsos}$ is implemented by a fully connected layer. 
The function $f_\mathrm{dinit}$ is also implemented by a fully connected layer.
The output vector of each LSTM block is fed to 5 functions $f_{t_u}$, $f_{t_v}$, $f_{L_u}$, $f_{L_e}$, and $f_{L_v}$ implemented by a fully connected layer. 
The 5 vectors output from these 5 functions are respectively converted to probability distributions $\boldsymbol{\xi}_{t_u}$, $\boldsymbol{\xi}_{t_v}$, $\boldsymbol{\xi}_{L_u}$, $\boldsymbol{\xi}_{L_e}$, and $\boldsymbol{\xi}_{L_v}$ through a softmax function, and these distributions predict one-hot vectors of 5-tuples in $\boldsymbol{s}_{j}$. 
In the learning process, the stacked LSTM and the fully connected layers are trained to predict $\boldsymbol{s}_{j+1}$ by a concatenated vector $\tilde{\boldsymbol{s}}_i=[\boldsymbol{\xi}_{t_u}, \boldsymbol{\xi}_{t_v}, \boldsymbol{\xi}_{L_u}, \boldsymbol{\xi}_{L_e}, \boldsymbol{\xi}_{L_v}]$ (See Section~\ref{ssec:Training} for detailes).
Summarizing the above, the process of the decoder part of our model is as follows: 
\begin{align}
\mathrm{SOS}&=f_\mathrm{dsos}([\boldsymbol{z}^T, \boldsymbol{C}_i^T]^T), \\  
\boldsymbol{h}_0&=f_{\rm dec}(f_\mathrm{dinit}(\boldsymbol{C}_i), [f_{\rm demb}(\mathrm{SOS})^T, \boldsymbol{z}^T, \boldsymbol{C}_i^T]^T), \label{eq:hidden_cond}\\
\boldsymbol{h}_{j+1}&=f_{\rm dec}(\boldsymbol{h}_{j}, [f_{\rm demb}(\boldsymbol{s}_j)^T, \boldsymbol{z}^T, \boldsymbol{C}_i^T]^T), \label{eq:dec_cond}\\
\boldsymbol{\xi}_{t_u} &= {\rm Softmax}(f_{t_u}(\boldsymbol{h}_j)), \\
% \end{align}
% \begin{align}
\boldsymbol{\xi}_{t_v} &= {\rm Softmax}(f_{t_v}(\boldsymbol{h}_j)), \\
\boldsymbol{\xi}_{L_u} &= {\rm Softmax}(f_{L_u}(\boldsymbol{h}_j)), \\
\boldsymbol{\xi}_{L_e} &= {\rm Softmax}(f_{L_e}(\boldsymbol{h}_j)), \\
\boldsymbol{\xi}_{t_v} &= {\rm Softmax}(f_{L_v}(\boldsymbol{h}_j)), \\
\tilde{\boldsymbol{s}}_j &= [\boldsymbol{\xi}_{t_u}^T, \boldsymbol{\xi}_{t_v}^T, \boldsymbol{\xi}_{L_u}^T, \boldsymbol{\xi}_{L_e}^T, \boldsymbol{\xi}_{L_v}^T]^T. 
\end{align}

%{\bf{Decoder}} DecoderはLSTMを用いており, 潜在ベクトル$\boldsymbol{z}$と条件ベクトル$c_i$を最初の入力とし, その二つのベクトルの値に応じた, 特徴のシーケンスを生成する. 
%提案モデルのDecoderは, シーケンスの学習を行う$f_{\rm dec}$, 埋め込み関数$f_{\rm emb}$, 潜在ベクトル$\boldsymbol{z}$を所定のサイズに変更する関数$f_\boldsymbol{z}$, 隠れ層の出力をそれぞれのシーケンスの5-tupleの要素$\boldsymbol{s}_j=(t_u, t_v, L_u, L_e, L_v)$に変換する関数で構成される. 
%$t_u, t_v, L_u, L_e, L_v$はそれぞれ, $\rm Softmax$関数によって表されたカテゴリカル分布からのサンプリングしたonehot vectorを表す. 
%$\boldsymbol{s}_j$は, onehot vector同士をvstackしたものとなっており, 5-tupleを表す. 
%$\boldsymbol{s}_0$は, シーケンスのSOSを表現するベクトルとなる. 
%学習時に, デコーダに対して$\boldsymbol{s}_{j-1}, c_i$を入力することで, 現在のシーケンスが条件$c_i$のシーケンスであることを学習させる. 
%生成時には, $c_i$を元に, $c_i$に従った特徴のシーケンスを生成させる. 

\subsection{Training}\label{ssec:Training}

Using sequence data fed by DFS code and structural feature vectors calculated by a statistical process, GraphTune infers the functions mentioned in Section~\ref{ssec:ModelArchitecture}.
In the training process, we input sequence data ${\boldsymbol{S}_i\in\mathcal{S}}$ and a condition vector $\boldsymbol{C}_i\in \mathcal{C}$ into the proposed model, and obtain a latent vector $\boldsymbol{z}$ and a predicted sequence $\tilde{\boldsymbol{S}}_i=\{\tilde{\boldsymbol{s}}_j | j=0, 1, \cdots , k\}$. 

Following the optimization manner of VAE~\cite{Kingma2014}, our model with encoder~$q_{\boldsymbol{\phi}}(\boldsymbol{z}|\boldsymbol{S}_i)$ and decoder~$p_{\boldsymbol{\theta}}(\boldsymbol{S}_i|\boldsymbol{z})$ considers the two components 
\begin{align}
  % \log p(\boldsymbol{S}_i) \ge -\mathbb{D}_\mathrm{KL}(q_{\boldsymbol{\phi}}(\boldsymbol{z}|\boldsymbol{S}_i)||p(\boldsymbol{z}))+\mathbb{E}_{q_{\boldsymbol{\phi}}(\boldsymbol{z}|\boldsymbol{S}_i)}[p_{\boldsymbol{\theta}}(\boldsymbol{S}_i|\boldsymbol{z})], \label{eq:lower_bound}
  -\mathbb{D}_\mathrm{KL}(q_{\boldsymbol{\phi}}(\boldsymbol{z}|\boldsymbol{S}_i)||p(\boldsymbol{z}))\quad\mathrm{and}\label{eq:lower_bound_1}\\
  \mathbb{E}_{q_{\boldsymbol{\phi}}(\boldsymbol{z}|\boldsymbol{S}_i)}[p_{\boldsymbol{\theta}}(\boldsymbol{S}_i|\boldsymbol{z})] \label{eq:lower_bound_2}
\end{align}
of the variational lower-bound, where $\mathbb{D}_\mathrm{KL}(\cdot)$ and $p(\boldsymbol{z})$ denote the Kullback-Leibler divergence and the multidimensional standard normal distribution~$\mathcal{N}(\boldsymbol{0}, \boldsymbol{I})$ with dimension $L$, respectively. 
Like the normal VAE~\cite{Kingma2014}, the first component Eq.~(\ref{eq:lower_bound_1}) regularizes the latent state distribution to be the standard normal distribution, and can be written as
\begin{align}
	{\rm Loss_{enc}}(\boldsymbol{\mu}, \boldsymbol{\sigma})=\frac{1}{2}\sum^L_{l=1}(1+\log((\sigma_l)^2) -(\mu_l)^2 - (\sigma_l)^2). 
\end{align}
The second component Eq.~(\ref{eq:lower_bound_2}) is a reconstruction loss that ensures the predicted sequence is similar to the input sequence from the dataset.
For our model, the reconstruction loss is defined for a sequence data $\boldsymbol{S}_i$ and a predicted sequence $\tilde{\boldsymbol{S}}_i$ by
\begin{align}
	{\rm Loss_{dec}}(\tilde{\boldsymbol{S}}_j, \boldsymbol{S}_j)=-\frac{1}{|\boldsymbol{S}|}\sum_{i=0}^{|\boldsymbol{S}|}\sum_{c}^{} \boldsymbol{s}_j(c)\log(\tilde{\boldsymbol{s}}_j(c)), 
\end{align}
where $\boldsymbol{s}(c)$ and $\tilde{\boldsymbol{s}}(c)$ represent component $c\in\{t_u,t_v , L_u, L_e , L_v \}$ of $s_i$ and $\tilde{\boldsymbol{s}}_i$, respectively.

By uniting the two losses with the idea of $\beta$-VAE~\cite{Higgins2017}, the proposed model is optimized by gradient descent on the following loss with weight $\beta$.
\begin{align}
  \mathrm{Loss}(\boldsymbol{\mu}, \boldsymbol{\sigma}, \tilde{\boldsymbol{s}}_j, \boldsymbol{s}_j)=\beta\cdot{\rm Loss_{enc}}(\boldsymbol{\mu}, \boldsymbol{\sigma})+{\rm Loss_{dec}}(\tilde{\boldsymbol{s}}_j, \boldsymbol{s}_j). 
\end{align}
The loss is backpropagated to the model, and we use the reparameterization trick~\cite{Kingma2014} for backpropagation through the Gaussian latent variable. 

The detailed algorithm of the training is shown in Algorithm~\ref{alg:Training}. 
For a given sequence dataset $\mathcal{S}=\{\boldsymbol{S}_1, \boldsymbol{S}_2, \dots \}$ and the condition vector set $\mathcal{C}=\{\boldsymbol{C}_1, \boldsymbol{C}_2, \dots\}$ corresponding to the dataset, Algorithm~\ref{alg:Training} returns learned encoder functions ($f_\mathrm{eemb}$, $f_\mathrm{enc}$, $f_{\boldsymbol{\mu}}$, and $f_{\boldsymbol{\sigma}^2}$) and decoder functions ($f_\mathrm{dinit}$, $f_\mathrm{demb}$, $f_\mathrm{dec}$, $f_{t_v}$, $f_{t_v}$, $f_{L_u}$, $f_{L_e}$, and $f_{L_v}$). 
First, the total loss is initialized (Line~2). 
The algorithm then iterates over all sequences $\boldsymbol{S}_i$ (Lines~3-23). 
An encoder recursively calculates $\boldsymbol{h}_j$ and a latent vector $\boldsymbol{z}$ is sampled (Lines~4-9). 
The loss for regularization of the latent state distribution is added to the total loss (Line~10). 
A predicted sequence $\tilde{\boldsymbol{S}}_i$, $\mathrm{SOS}$ and $\boldsymbol{h}_0$, are initialized by an empty vector, $f_\mathrm{dsos}$, and $f_\mathrm{dinit}$, respectively (Line~11-13). 
$\boldsymbol{h}_j$ is converted to probability distributions of one-hot vectors of 5-tuples through the function $f_{t_v}$, $f_{t_v}$, $f_{L_u}$, $f_{L_e}$, and $f_{L_v}$ (Lines~15-17). 
Next, the distribution is vertically concatenated into a single vector $\tilde{\boldsymbol{s}}_j$, and these vectors $\{\tilde{\boldsymbol{s}}_j| j=1,2,\dots , |\boldsymbol{S}_i|\}$ are horizontally concatenated into a predicted sequence $\tilde{\boldsymbol{S}}_i$ (Lines~18-19).
A decoder also recursively calculates $\boldsymbol{h}_j$ for the prediction of the next 5-tuple (Line~20).
The reconstruction loss calculated from the concatenated probability distributions is added to the total loss (Line~22). 
Lastly, the weights of all functions are updated by back-propagating the total loss (Line~24). 
The above procedures are iterated until the total loss converges (Line~25).

\begin{algorithm}[tb]
   \caption{Training of GraphTune}         
  \label{alg:Training}
  \begin{algorithmic}[1]
    \REQUIRE{Graph dataset $\mathcal{S}=\{\boldsymbol{S}_1, \boldsymbol{S}_2, \dots \}$, \\
    Condition vector set $\mathcal{C}=\{\boldsymbol{C}_1, \boldsymbol{C}_2, \dots\}$}
    \ENSURE{Learned functions $f_\mathrm{eemb}$, $f_\mathrm{enc}$, $f_{\boldsymbol{\mu}}$, $f_{\boldsymbol{\sigma}^2}$, $f_\mathrm{dinit}$, $f_\mathrm{demb}$, $f_\mathrm{dec}$, $f_{t_u}$, $f_{t_v}$, $f_{L_u}$, $f_{L_e}$, and $f_{L_v}$}
    \REPEAT
        \STATE $\mathrm{Loss} \leftarrow 0$
        \FOR{$i$ from 1 to $|\mathcal{S}|$}
        % encoder step
        \STATE $\boldsymbol{h}_0 \leftarrow \boldsymbol{0}$
        \FOR{$j$ from 0 to $k-1$}
          \STATE $\boldsymbol{h}_{j+1} \leftarrow f_{\rm enc}(\boldsymbol{h}_{j}, f_{\rm eemb}([\boldsymbol{s}_{j}^T, \boldsymbol{C}_i^T]^T))$
        \ENDFOR
        \STATE $\boldsymbol{\mu} \leftarrow f_{\boldsymbol{\mu}}(\boldsymbol{h}_k)$; $\boldsymbol{\sigma}^2 \leftarrow f_{\boldsymbol{\sigma}^2}(\boldsymbol{h}_k)$
        \STATE $\boldsymbol{z} \sim \mathcal{N}(\boldsymbol{\mu}, \boldsymbol{\sigma}^2)$
        \STATE $\mathrm{Loss} \leftarrow \mathrm{Loss}+\beta\cdot{\rm Loss_{enc}}(\boldsymbol{\mu}, \boldsymbol{\sigma})$
  
        % decoder step
        \STATE $\tilde{\boldsymbol{S}}_i\leftarrow []$
        \STATE $\mathrm{SOS}\leftarrow f_\mathrm{dsos}([\boldsymbol{z}^T, \boldsymbol{C}_i^T]^T)$
        \STATE $\boldsymbol{h}_0 \leftarrow f_{\rm dec}(f_\mathrm{dinit}(\boldsymbol{C}_i), [f_{\rm demb}(\mathrm{SOS})^T, \boldsymbol{z}^T, \boldsymbol{C}_i^T]^T)$
        \FOR{$j$ from 0 to $k-1$}
          \FOR{$c\in (t_u, t_v, L_u, L_e, L_v)$}
            \STATE $\boldsymbol{\xi}_c \leftarrow {\rm Softmax}(f_c(\boldsymbol{h}_j))$
          \ENDFOR
          \STATE $\tilde{\boldsymbol{s}}_j \leftarrow [\boldsymbol{\xi}_{t_u}^T, \boldsymbol{\xi}_{t_v}^T, \boldsymbol{\xi}_{L_u}^T, \boldsymbol{\xi}_{L_e}^T, \boldsymbol{\xi}_{L_v}^T]^T$
          \STATE $\tilde{\boldsymbol{S}}_i\leftarrow [\tilde{\boldsymbol{S}}_i, \tilde{\boldsymbol{s}}_j]$
          \STATE $\boldsymbol{h}_{j+1} \leftarrow f_{\rm dec}(\boldsymbol{h}_{j}, 
          [f_{\rm demb}(\boldsymbol{s}_j)^T, \boldsymbol{z}^T, \boldsymbol{C}_i^T]^T)$
        \ENDFOR
        \STATE $\mathrm{Loss} \leftarrow \mathrm{Loss}+{\rm Loss_{dec}}(\tilde{\boldsymbol{S}}_i, \boldsymbol{S}_i)$
        \ENDFOR
      \STATE Back-propagate $\mathrm{Loss}$ and upate weights
    \UNTIL{stopping criteria}
  \end{algorithmic}
\end{algorithm}

%推論時, 提案法の学習はAlgorithm \ref{alg1}に示す手順で行う. 
%まず, グラフのデータセットを$\bm{G}$, 最小DFSコードによりシーケンスデータセット$\mathcal{S}$に変換する(1行目). 
%その後, 全ての学習されていないニューラルネットのパラメータの初期化を行う(2行目). 
%シーケンスの各要素$\boldsymbol{s}_{j-1}$に対して$c_i$をvstackしたものを$f_{\rm emb}$によって埋め込んだベクトルと, 前回のステップでのLSTMの出力$\boldsymbol{h}_{j-1}$を再帰的に, $f_{\rm enc}$に入力する(8-10行目). 
%シーケンス$S$を, Encoderに対して入力し終えたのち, 最後の出力である$\boldsymbol{h}_k$を, $f_{\mu}, f_{\sigma}$に入力することで, パラメータ$\mu, \sigma$へ変換する(11行目). 
%reparameterization trickにより, $\mu, \sigma$を用いて, 潜在変数$\boldsymbol{z}$をサンプリングする(12行目). 
%$\rm Loss_{enc}$により, $\mu, \sigma$より, Encoderの誤差を計算し, 変数$Loss$に加算する(13行目). 
%ベクトル$\mu, \sigma$の次元が$L$である時, $Loss_{\rm enc}$は次式で表される. 
%$\mu_i, \sigma_i$は, ベクトル$\mu, \sigma$の$i$次元目の要素を表す.

% %$f_\boldsymbol{z}$により, 潜在変数$\boldsymbol{z}$をDecoderへの最初の入力(SOS)である$\boldsymbol{s}_0$に変換する(14行目). 
% %Encoderと同様に, シーケンスの各要素$\boldsymbol{s}_{j-1}$に対して$c_i$をvstackしたベクトルを$f_{\rm emb}$によって埋め込んだベクトルと, 前回のステップでのLSTMの出力$\boldsymbol{h}_{j-1}$を再帰的に, $f_{\rm dec}$に入力する(17-18行目). 
% %LSTMの出力$\boldsymbol{h}_j$は, 5つの線形層により,  $c\in \{t_u, t_v, L_u, L_e, L_v\}$の予測分布$\boldsymbol{\xi}_c$をそれぞれ計算し, 全てをvstackしたものを$\tilde{\boldsymbol{s}}_j$とする(19-23行目). 
% %その後, Decoderの誤差を, $\rm Loss_{dec}$を用いて計算し, 変数$Loss$に加算する(24行目). 
% %誤差関数$\rm Loss_{dec}$を次式に示す. 
% %$s(c)$は, ベクトル$s$における, $c\in \{t_u, t_v, L_u, L_e, L_v\}$に対応した次元を抜き出す操作に当たる. 
% %また$\rm \log$はベクトルの各要素に対して, 計算を行う. 
% \begin{eqnarray}
% 	{\rm Critrion_{dec}}(\tilde{\boldsymbol{s}}_j, \boldsymbol{s}_j)=-\sum_{c}^{} \boldsymbol{s}_j(c)\log(\tilde{\boldsymbol{s}}_j(c))
% \end{eqnarray}
% %計算した$Loss$を用いて, 勾配情報の更新を行う(28行目). 

\subsection{Generation}

When we generate graphs with specific structural features, GraphTune recursively generates sequential data in the DFS code format using learned functions.
% アルゴリズムを参照しながら説明を書く．
We give a sampled latent vector and a condition vector whose elements are tuned to specific values to the decoder. 
By giving a condition vector, the decoder recursively generates a sequence of 5-tuples according to the condition vector. 
Finally, we get a graph with specific structural features by inverse converting from sequence data to a graph.

The entire procedure for the generation of a graph with a specific condition is summarized in Algorithm~\ref{alg:Generation}. 
The input and output of the algorithm are a condition vector $\boldsymbol{C}$ with specific values and sequence data of a graph with the condition, respectively.
Firstly, a generated sequence data $\boldsymbol{S}$ and an iterator variable $j$ are initialized (Lines~1-2). 
For the generation, a latent vector $\boldsymbol{z}$ is sampled from the standard normal distribution $\mathcal{N}(\boldsymbol{0}, \boldsymbol{I}^2)$ (Line~3). 
$\boldsymbol{h}_0$ is calculated from the sampled latent vector $\boldsymbol{z}$ and the given condition $\boldsymbol{C}$ (Lines~4-5).
To obtain the next 5-tuple in a predicted sequence, the element-wise distributions $\xi_{t_u}$, $\xi_{t_v}$, $\xi_{L_u}$, $\xi_{L_e}$, and $\xi_{L_v}$ of 5-tuple are calculated, and predicted values $\hat{\boldsymbol{s}}(t_u)$, $\hat{\boldsymbol{s}}(t_v)$, $\hat{\boldsymbol{s}}(L_u)$, $\hat{\boldsymbol{s}}(L_e)$, and $\hat{\boldsymbol{s}}(L_v)$ are sampled from these distributions, respectively (Lines~7-10). 
The predicted values $\hat{\boldsymbol{s}}(t_u)$, $\hat{\boldsymbol{s}}(t_v)$, $\hat{\boldsymbol{s}}(L_u)$, $\hat{\boldsymbol{s}}(L_e)$, and $\hat{\boldsymbol{s}}(L_v)$ of each element of 5-tuple are vertically concatenated into a single vector $\hat{\boldsymbol{s}}_j$, and the concatenated vector $\hat{\boldsymbol{s}}_j$ is horizontally concatenated to the predicted sequence $\boldsymbol{S}$ (Lines~11-12). 
$\boldsymbol{h}_j$ is recursively calculated for the prediction of the next 5-tuple, and the iterator variable $j$ is updated (Lines~13-14). 
To stop the generation of the sequence in finite size, the iteration finishes if at least one element of the prediction $\hat{\boldsymbol{s}}_j$ of 5-tuple is EOS (Line~15).
A graph with a specific condition $\boldsymbol{C}$ is easily constructed from the sequence $\boldsymbol{S}$ at the end of the algorithm. 

Most parts of the algorithm of GraphTune are deterministic, but the latent vector $\boldsymbol{z}$ and the 5-tuple in DFS code are randomly sampled from estimated distributions. 
By the sampling process, GraphTune is possible to generate a wide variety of graphs under a single condition. 
Based on the basic concept of VAE, the final output of the encoder of GraphTune is not the latent vector $\boldsymbol{z}$, but the parameters $\boldsymbol{\mu}$ and $\boldsymbol{\sigma}$ of the normal distribution over the latent space. 
The latent vector $\boldsymbol{z}$ is sampled from the normal distribution specified by the parameters.
Similarly, the decoder output in GraphTune is a distribution from which 5-tuples in DFS code will be sampled. 

% ■■■ 生成のランダム性がどこに由来するか書く ■■■  済
% ToDo: f_dsosとf_dembは数式上一つにまとめて，2レイヤのFC層と書く．

%生成時はAlgoritm \ref{alg2}に示すように, 生成した5-tupleを再帰的に入力し, 生成を行う.
%まず入力として, 生成したい特徴量の値を要素としたベクトルである$C$を受け取る. 
%標準正規分布$N$より, 潜在変数$\boldsymbol{z}$をサンプリングする(2行目). 
%潜在変数$\boldsymbol{z}$を関数$f_\boldsymbol{z}$によって, Decoderへの最初の入力である$\boldsymbol{s}_0$に変換する(4行目). 
%前回のステップで生成したシーケンスの要素$\tilde{\boldsymbol{s}}_{i-1}$に対して$C$をvstackしたものを$f_{\rm emb}$によって埋め込んだベクトルと, 前回のステップでのLSTMの出力$\boldsymbol{h}_{i-1}$を再帰的に, $f_{dec}$に入力し, 現在のステップの出力である$\boldsymbol{h}_i$を得る(7行目). 
%LSTMの出力$\boldsymbol{h}_i$は, 15-19式に示す5種類の線形層により,  $c\in \{t_u, t_v, L_u, L_e, L_v\}$の予測分布$\boldsymbol{\xi}_c$をそれぞれ計算し, 全てをvstackしたものを$\tilde{\boldsymbol{s}}_i$とする(10-12行目). 
%シーケンスのリストに対して, ${\rm hstack}$を用いて新たな5-tupleを追加する.(14行目) 
%この生成プロセスをEOSが生成されるまで繰り返す(16行目).

In the generating process of a graph from a DFS code, we construct the graph by creating nodes and edges based on the timestamps of the nodes in the DFS code that is output by the decoder. 
Since this research focuses on generating unlabeled graphs, we ignore label information in the DFS code. 
Although GraphTune attempts to faithfully reproduce the DFS code in the graphs of the dataset, GraphTune can not guarantee full compliance with the rules of DFS code.
Therefore, 5-tuples that violate the rule of DFS code may be output. 
A typical case is that a 5-tuple contained in the DFS code reappears in subsequent DFS code. 
To solve these problems, we ignore any 5-tuple that conflicts with the 5-tuples that appear in the DFS code before that.
% ■■■ DFSコードから戻す際の手続きを書く．矛盾するDFSコードが出ることがあると書く．■■■  済

\begin{algorithm}                      
	\caption{Generation of a graph with a specific condition}         
	\label{alg:Generation}
	\begin{algorithmic}[1]
    \REQUIRE{Condition vector $\boldsymbol{C}$ with specific values}
    \ENSURE{Sequence data $\boldsymbol{S}$ of a graph $G$ with a specific condition $\boldsymbol{C}$}
		\STATE $\boldsymbol{S}\leftarrow []$
		\STATE $j\leftarrow 0$
		\STATE $\boldsymbol{z}\sim \mathcal{N}(\boldsymbol{0}, \boldsymbol{I}^2)$
    \STATE $\mathrm{SOS}\leftarrow f_\mathrm{dsos}([\boldsymbol{z}^T, \boldsymbol{C}^T]^T)$
		\STATE $\boldsymbol{h}_0 \leftarrow f_{\rm dec}(f_\mathrm{dinit}(\boldsymbol{C}), [f_{\rm demb}(\mathrm{SOS})^T, \boldsymbol{z}^T, \boldsymbol{C}^T]^T)$
		\REPEAT
      \FOR{$c\in (t_u, t_v, L_u, L_e, L_v)$}
		    \STATE $\boldsymbol{\xi}_{c} \leftarrow {\rm Softmax}(f_{c}(\boldsymbol{h}_j))$
		    \STATE $\hat{\boldsymbol{s}}(c)\sim \boldsymbol{\xi}_c$
		  \ENDFOR
      \STATE $\hat{\boldsymbol{s}}_j\leftarrow [\hat{\boldsymbol{s}}(t_u)^T, \hat{\boldsymbol{s}}(t_v)^T, \hat{\boldsymbol{s}}(L_u)^T, \hat{\boldsymbol{s}}(L_e)^T, \hat{\boldsymbol{s}}(L_v)^T]^T$
		  \STATE $\boldsymbol{S}\leftarrow [\boldsymbol{S}, \hat{s}_j]$
		  \STATE $\boldsymbol{h}_{j+1} \leftarrow f_{\rm dec}(\boldsymbol{h}_{j}, [f_{\rm demb}(\hat{\boldsymbol{s}}_j)^T, \boldsymbol{z}^T, \boldsymbol{C}^T]^T)$
		  \STATE $j\leftarrow j+1$
		\UNTIL{$\exists c: \hat{\boldsymbol{s}}(c)={\rm EOS}$}
	\end{algorithmic}
\end{algorithm}

\subsection{Necessity of CVAE Architecture}

The most distinctive point of GraphTune is that it allows continuous tuning of a structural feature of generated graphs by giving condition vectors to the decoder and the encoder in VAE, respectively.
Although the architecture of GraphTune is similar to that of GraphGen, GraphTune differs from GraphGen in that it is divided into an encoder and a decoder.
Our goal is to tune only the feature that is specified in the condition vector while maintaining the values of the other features.
In order to achieve this goal, it is important not only to simply learn graph features, but also to separate the features into the feature specified in the condition vector (conditioned features) and the other features (unconditioned features). 
Since the decoder of GraphTune is given a condition vector along with a latent vector, it can proceed with learning to reduce the reconstruction loss based on accurate information regarding conditioned features.
Therefore, the latent vector does not have to contain information about conditioned features. 
On the contrary, it is desirable that the latent vector contains only information about unconditioned features.
In GraphTune, by explicitly giving the encoder a condition vector as well as the decoder, the encoder can remove information about the conditioned features from the latent vector.

The architecture of GraphTune, which provides condition vectors to both the encoder and the decoder in VAE, is a reasonable architecture for accurately tuning features of generated graphs.
% ■■■ 追加するか検討 ■■■ One of the simplest implementations of providing a condition vector in graph generation is to concatenate a condition vector to the input vectors of GraphGen. 
% ■■■ 追加するか検討 ■■■ % Note that the architecture of GraphGen is close to the decoder part from GraphTune.
% ■■■ 追加するか検討 ■■■ However, when actually implemented, such a simple architecture cannot demonstrate sufficient tunability. 
% ■■■ 追加するか検討 ■■■ Similarly, an architecture that inputs a condition vector to either the encoder or the decoder does not have significant tunability.
An architecture that inputs a condition vector to either the encoder or the decoder does not have significant tunability.
The detailed experimental results that support the necessity of the CVAE-based architecture of GraphTune are presented in Section~\ref{ssec:AblationExperiment}. 
% ■■■ Latent vector はcondiiton以外の情報を埋め込んだベクトル．Decoder 側は正確なconditionを知っているので，embedからはconditionの情報は取り除かれる ■■■  済
% ■■■ GraphTuneはGraphGenに似た構造をしているが，encoderとdecoderに分けて実装している点が異なり，これは重要な意味を持つ．条件付き生成の最もシンプルな実装は GraphGen にcondition vectorを与えること．でも実際にやってみるとうまく行かない．■■■  済

\section{Experiments}\label{sec:Experiments}

We verify that GraphTune can learn structural features from graph data and generate a graph with specific structural features. 
In this section, we first present performance evaluations of GraphTune on a graph dataset generated with WS model~\cite{Watts1998}, which is representative of statistical graph generation models.
In addition, we also present a performance evaluation of GraphTune on a real graph dataset extracted from a who-follows-whom network of Twitter.
% ■■■ WS と BA もやったことも書く ■■■  済
Through the evaluations, we show that GraphTune yields better performance than the conventional generative models, namely, GraphGen and CondGen.
% Twitterをつかったこと，CondGen，GraphGenと比較したことなどを簡単に書く．

\subsection{Baselines}\label{ssec:Baselines}

To confirm the basic characteristics of GraphTune in a conditional graph generation task, we compare the performance of GraphTune with two baseline models: GraphGen~\cite{Goyal2020} and CondGen~\cite{Yang2019}. 

\subsubsection*{GraphGen}
GraphGen employs a scalable approach to domain-agnostic labeled graph generation and is a representative model that adopts a sequence data-based approach. 
As we mentioned in the introduction, the sequence data-based approach is one of the most successful approaches in the field of learning-based graph generation.
GraphGen was compared with DeepGMG~\cite{Li2018} and GraphRNN~\cite{You2018} in~\cite{Goyal2020}. 
It was reported that GraphGen is superior to these methods in terms of the reproduction accuracy of graph structural features. 
Although GraphGen is an outstanding model, it unfortunately does not provide conditional generation of graphs.
Hence, GraphGen is a baseline in terms of the reproduction accuracy of graph structural features, and it does not provide a baseline regarding the ability of conditional generation.
In the evaluations in Section~\ref{sec:Experiments}, we use the parameters recommended in~\cite{Goyal2020}. 
% GraphGenはシーケンスベースの生成を使う方法で成功した手法なので選んだ．
% 著者が公開するソースコードを使って推奨のパラメータを使ったと書く．

\subsubsection*{CondGen}
CondGen employs conditional structure generation through graph variational generative adversarial nets and is one of the few models that achieves a conditional generation for general graphs that is not limited to a specific domain.
To the best of our knowledge, CondGen is almost the only model that is oriented towards the reproduction of global-level structural features in human relationship graphs including social networks and citation networks.
GraphGen was compared with GraphVAE~\cite{Simonovsky2018}, NetGAN~\cite{Bojchevski2018}, and GraphRNN~\cite{You2018} in ~\cite{Yang2019}. 
The study~\cite{Yang2019} reports that CondGen records the best performance in most cases. 
Since CondGen supports a conditional generation of graphs, CondGen provides a baseline regarding the ability of conditional generation.
Unlike GraphTune specifying a value of a feature after a training process, CondGen requires training datasets grouped by labels. 
Hence, when we specify another condition, CondGen needs to relearn another dataset grouped by the conditions. 
In the evaluations in Section~\ref{sec:Experiments}, we use the parameters recommended in the paper~\cite{Yang2019}. 
Since the number of nodes and edges are required as input parameters in the generation, the number of nodes and edges of sampled graphs with a specific label from the dataset are used. 
% global-level featureとしてどんなものを再現しているか，データセットとしてどんなものを使っているか書く．
% CondGenは数少ないドメインフリーかつ条件付き生成を可能にしたモデルなので選んだということを書く．しかも，Global-level feature を調整可能である．著者が公開するソースコードを使って著者推奨のパラメータを使用と書く．
% グループ化する手法なので，指定した以外の種類を生成することはできない．

\subsection{Parameters and Training Dataset}\label{ssec:ParametersandTrainingDataset}

The parameters of a model in GraphTune are set as follows.
For the encoder part of our model, we use 2-layer LSTM blocks for $f_\mathrm{enc}$, which has a hidden state vector of dimension 223.
% We use a dropout of probalility 0 in the LSTM layers of the encoder part.
The dimension of $f_\mathrm{eemb}$ is set to 227. 
The dimensions of vectors $\boldsymbol{\mu}$ and $\boldsymbol{\sigma}^2$, that is, the dimension of the latent vector $\boldsymbol{z}$, are set to 10. 
Three-layer LSTM blocks for $f_\mathrm{dec}$ which have a hidden state vector of dimension 250, is adopted for the decoder part. 
% We use a dropout of probalility 0 in the LSTM layers of the decoder part.
The dimensions of $f_\mathrm{dsos}$ and $f_\mathrm{demb}$ are set to 250. 
We use Adam optimizer to train the neural networks for 10000 epochs with a batch size of 37 and an initial learning rate of 0.001 for training.
The weight $\beta$ for the calculation of loss is set to 3.0. 
% The learning is regularized by a weight decay of XXX.
% GraphTuneのパラメータを説明する．

%提案法のモデルのパラメータは, Encoder, DecoderのLSTMの隠れ層の次元は$256$, 埋め込み層の次元は$128$, Encoderの正規分布のパラメータに変換を行う線形層は$20$次元である. 
%なお最適化アルゴリズムはAdamを用いており, 初期学習係数は$0.001$, weight decayは$0.01$, 重みのclip閾値は$0.015$である. 
%ミニバッチ数は$60$, エポック数は$400$である. 

To investigate the basic performance of GraphTune, we constructed a graph dataset (WS dataset) generated by WS model.
As the parameters of WS model, we set average degree $K$ to 3, and randomly change rewiring probability $p$ within the range $[0.1, 0.6]$.
As we mentioned in the introduction, since stochastic graph generation models generate graphs with simple algorithms, they focus on only a single-aspect feature. 
Therefore, in graph datasets that are generated by these models, a part of the features that are not focused on are almost invariable. 
In the case of WS model, since it is a model that aims to reproduce small worldness, the average shortest path length changes greatly, but the other features are almost invariable. 
Our WS dataset contains average shortest path lengths ranging from 5.32 to 17.0.
% ■■■ WSとBAの説明 人工的に生成したものなので，画一的でfeatureのバリエーションが少ない．一部の特性についてはほぼ同じ値． ■■■ ほぼ済 数値を代入する必要あり．

To evaluate the performance of GraphTune on real graph dataset, we sampled data from the Twitter who-follows-whom graph in the Higgs Twitter Dataset~\cite{Domenico2013}. 
To prepare a human relationship graph dataset with sufficient size for training and evaluation, we sampled 2000 graphs from a single huge graph included in the Higgs Twitter Dataset with 456,626 nodes and 14,855,842 edges.
A graph in the dataset for the evaluations is sampled by performing a random walk that starts from a randomly selected node.
An initial node of a random walk is selected following a uniform distribution.
The edge for the next hop is randomly selected from all edges with equal probability.
The random walk ends up after 50 nodes are found, and a graph in the evaluation dataset is an induced subgraph that is composed of the nodes included in the random walk. 
Note that an edge can be included in the evaluation dataset if both nodes are included in the random walk, even if the edge is not included in the random walk.
Although the original Higgs Twitter Dataset is a directed graph, we ignore the directions of all edges in this study since our model is designed for undirected graphs.
This dataset with small and uniform size graphs is suitable for evaluating the basic tunability of global-level features without being affected by the difficulty of reproducing heterogeneous or very large graphs.
We split the dataset into two parts: the training set and validation set.
The size ratio of the training set and validation set are 90\% and 10\%, respectively.

% データセットの性質やサンプリングの方法などを書く．無向グラフにしていることを書く．比較的小さなサイズのデータセットはないため，サンプリングにより作成した．

\subsection{Structural Features}\label{ssec:StructuralFeatures}

As structural features of graphs, we focus on the following 5 features: average of shortest path length, average degree, modularity~\cite{Newman2004}, clustering coefficient, and a power-law exponent of a degree distribution~\cite{Jeff2014}. 
The value of modularity is calculated for modules consisting of nodes divided by the Louvain algorithm~\cite{Blondel2008}. 
We calculate the power-law exponent of the degree distribution by the powerlaw Python package~\cite{Jeff2014}. 
These global-level structural features are selected from survey papers~\cite{Boccaletti2006,Costa2007} on the measurement of complex network structures, and they have been widely used as graph features of human relationship graphs~\cite{Kwak2010,Viswanath2009}. 
Compared with local structures such as the number of nodes and edges, these global-level structural features are difficult to tune by adding or removing local structure to or from a graph, such as a node, edge, or hexagon.
Needless to say, adding or removing local structures to or from a graph can change the value of global-level structural features.
However, if we control a value of a feature to a specific value on a graph, we must understand the structure of the whole graph and consider the effect of the local structure on the value of the feature. 
For the creation of the dataset, the value of features is rounded to one decimal place.
% なぜこのような統計をとったのかを説明する．高次元の統計量が入っていることを書く．modularityのアルゴリズムなどを説明する．

\subsection{Performance Evaluations}\label{ssec:PerformanceEvaluations}

In this section, we show that GraphTune can generate graphs with specific structural features.
Performance comparison among three methods (GraphTune, CondGen, and GraphGen) and detailed analysis of generated graphs are provided.

We trained GraphTune, CondGen, and GraphGen using the training set described in Section~\ref{ssec:ParametersandTrainingDataset}, and generated graphs with specific conditions. 
For GraphGen, a single model is trained with the training set, since GraphGen does not provide conditional generation of graphs.
For GraphTune and CondGen, the models were trained individually for each feature that is focused on; that is, we trained 5 different models for each single feature. 
After the training process, we generated 300 graphs for each model. 
For each feature, we picked up 3 typical values of a feature from the range of values of the training set as the values of the condition vectors.
In the generation process, we give the condition vectors to models of GraphTune. 
Since CondGen requires training sets grouped by labels, the training sets are divided into 3 groups at the middle of the typical values.
Note that we cannot give a condition to GraphGen since it is designed for unconditional generation.
% 評価結果を説明する．

The summary of the results of generation for WS dataset and Twitter dataset are listed in TABLE~\ref{tab:average_ws} and \ref{tab:average}, respectively. % ■■■ 別のテーブルも参照 ■■■  済
The values of features other than average shortest path length of graphs in WS dataset are almost invariant, we list only results regarding average shortest path length in TABLE~\ref{tab:average_ws}.
In TABLE~\ref{tab:average} regarding the results for Twitter dataset, we list the results for all 5 features shown in Section~\ref{ssec:StructuralFeatures}.
The values in the columns of GraphTune, CondGen, GraphGen represents average values of features in graphs generated by each model.
Since GraphGen does not provide conditional generation, the same value is listed for different conditions.
We can consider that a method has better performance if the average value is closer to the values of the condition in the tunability of a condition.
The best performance achieved under each condition for a particular feature is emphasized in bold font.

As shown in TABLE~\ref{tab:average_ws}, GraphTune well reproduces the shift of the average shortest path length with the change in rewiring probability on WS dataset.
The value of the average shortest path length of the graphs generated by GraphTune is close to the value specified by the condition vector.
The average shortest path length for CondGen cannot be calculated since most of generated graphs are unconnected.
Then, we depict ``--'' for unconnected graphs. 
For reference, the average shortest path length of the largest component is given in parenthesis for unconnected graphs.
In contrast, all graphs generated by GraphTune were connected graphs.
This is an advantage of the sequence data-based approach taken by GraphTune (and GraphGen).
Even comparing the values of the average shortest path length of the largest components, it can be confirmed that the tunability of GraphTune is higher than that of CondGen.
% From the distribution of the average shortest path length in the graph generated by GraphTune shown in Fig.~\ref{fig:dist}, we can confirm that the distribution shifts along with the value of the condition.

% \begin{figure}[tbp]
%   \begin{center}
%     \includegraphics[width=\linewidth]{pic/dummy.pdf}
%   \end{center}
%   \caption{
%     ■■■ キャプションを書く ■■■
%   }
%   \label{fig:dist}
% \end{figure}

% ■■■ WSとBAの結果の説明 ■■■  済
% ■■■ 可視化データの参照 ■■■  済

\begin{table*}[tb]
  \caption{
    The average shortest path length in graphs generated by GraphTune, CondGen, and GraphGen for WS dataset. 
    We present the same value for all results of GraphGen for each feature since GraphGen does not provide conditional generation of graphs.
    The best performance between GraphTune and CondGen achieved under each condition is highlighted in bold font. 
    }
  \label{tab:average_ws}
  \centering
   \begin{tabular}{rc|c|c|c|c|}
    Global-level structural feature & Condition & GraphTune & CondGen & GraphGen & WS data \\
    & & average & average & average & average \\
    & & & & & (25-percentile / median / 75-percentile) \\\hline\hline
    &  7.5 & \bf 10.8  & -- ( 2.28 ) &      &       \\\cline{2-4}Average of shortest path length 
    & 10.0 & \bf 14.5  & -- ( 2.27 ) & 9.51 & 9.10 \\\cline{2-4}
    & 12.5 & \bf 16.8  & -- ( 2.30 ) &      & (7.67 / 8.74 / 10.2)      \\\hline
    % &    &       &   &  &       \\\cline{2-4}Average degree 
    % &    &       &   &  &   \\\cline{2-4}
    % &    & \bf   &   &  & ( /  / )    \\\hline
    % &    & \bf   &   &  &       \\\cline{2-4}Modularity 
    % &    & \bf   &   &  &  \\\cline{2-4}
    % &    & \bf   &   &  & ( /  / )      \\\hline
    % &    & \bf   &   &  &       \\\cline{2-4}Clustering coefficient 
    % &    & \bf   &   &  &  \\\cline{2-4}
    % &    & \bf   &   &  & ( /  / )      \\\hline
    % &    & \bf   &   &  &       \\\cline{2-4}Power-law exponent of a degree distribution
    % &    & \bf   &   &  &   \\\cline{2-4}
    % &    & \bf   &   &  & ( /  / )      \\\hline
    % & 0.050 & \bf        &     0.0573 &        &       \\\cline{2-4}Edge density 
    % & 0.075 & \bf        &     0.0731 & 0.0623 & 0.0732\\\cline{2-4}
    % & 0.10  & \bf        &     0.105  &        &       \\\hline
    % & 10    &            &     --     &        &       \\\cline{2-4}Diameter
    % & 20    &            &     --     & 12.647 & 11.4  \\\cline{2-4}
    % & 30    &            &     --     &        &       \\\hline
   \end{tabular}
\end{table*}

\begin{figure*}[tb]
  \begin{center}
    \includegraphics[width=0.3\linewidth]{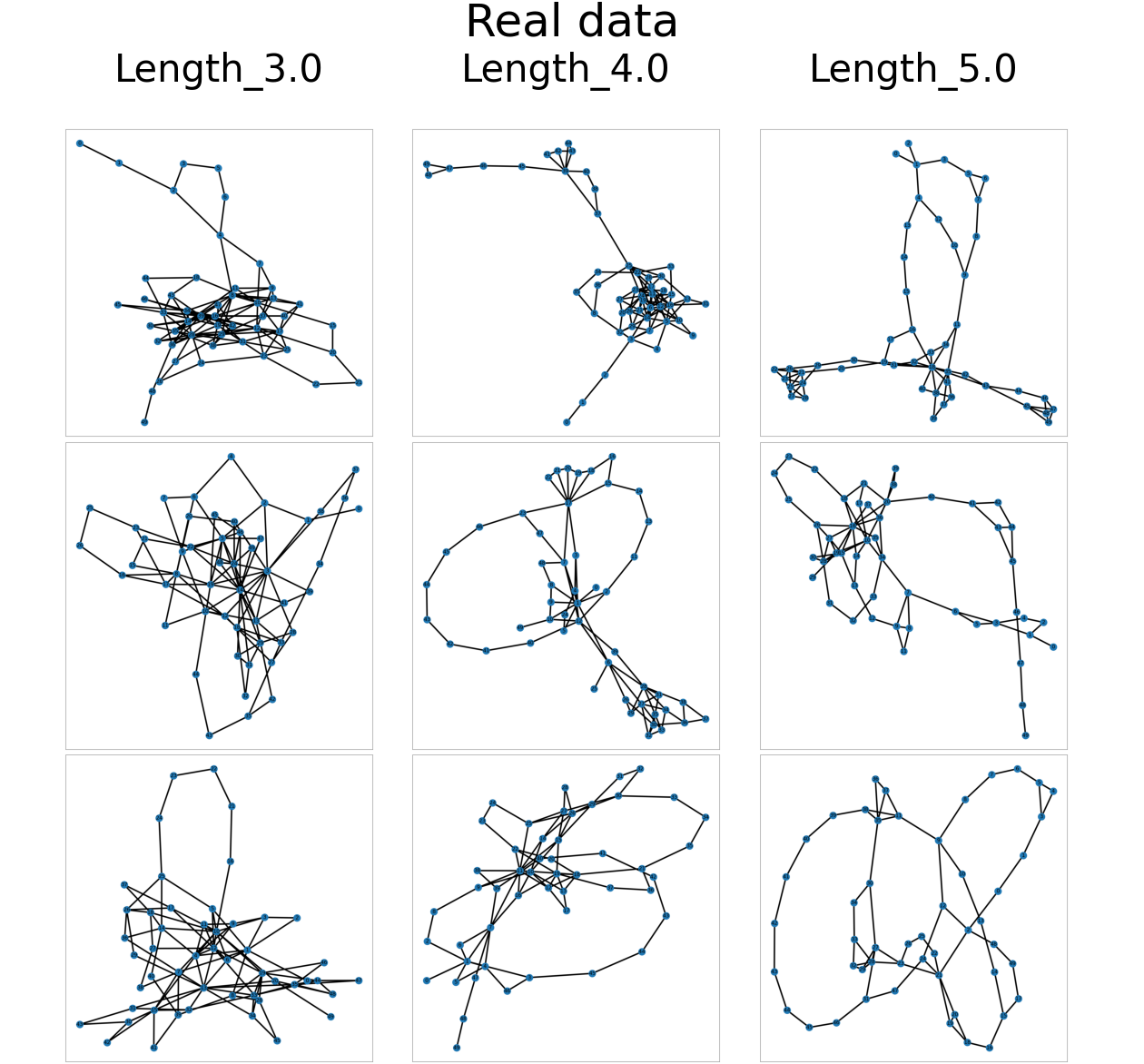}%
    \includegraphics[width=0.3\linewidth]{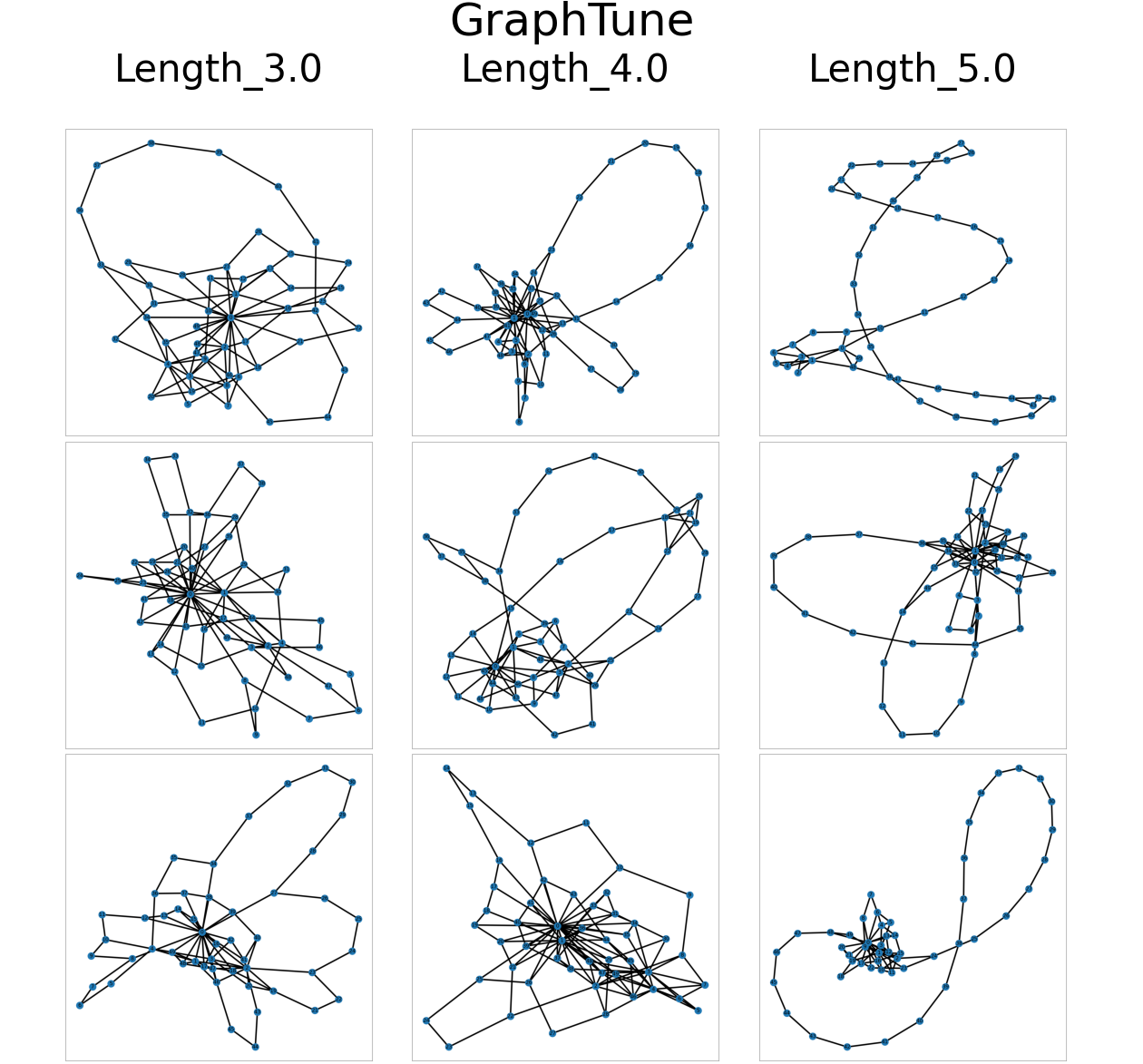}%
    \includegraphics[width=0.3\linewidth]{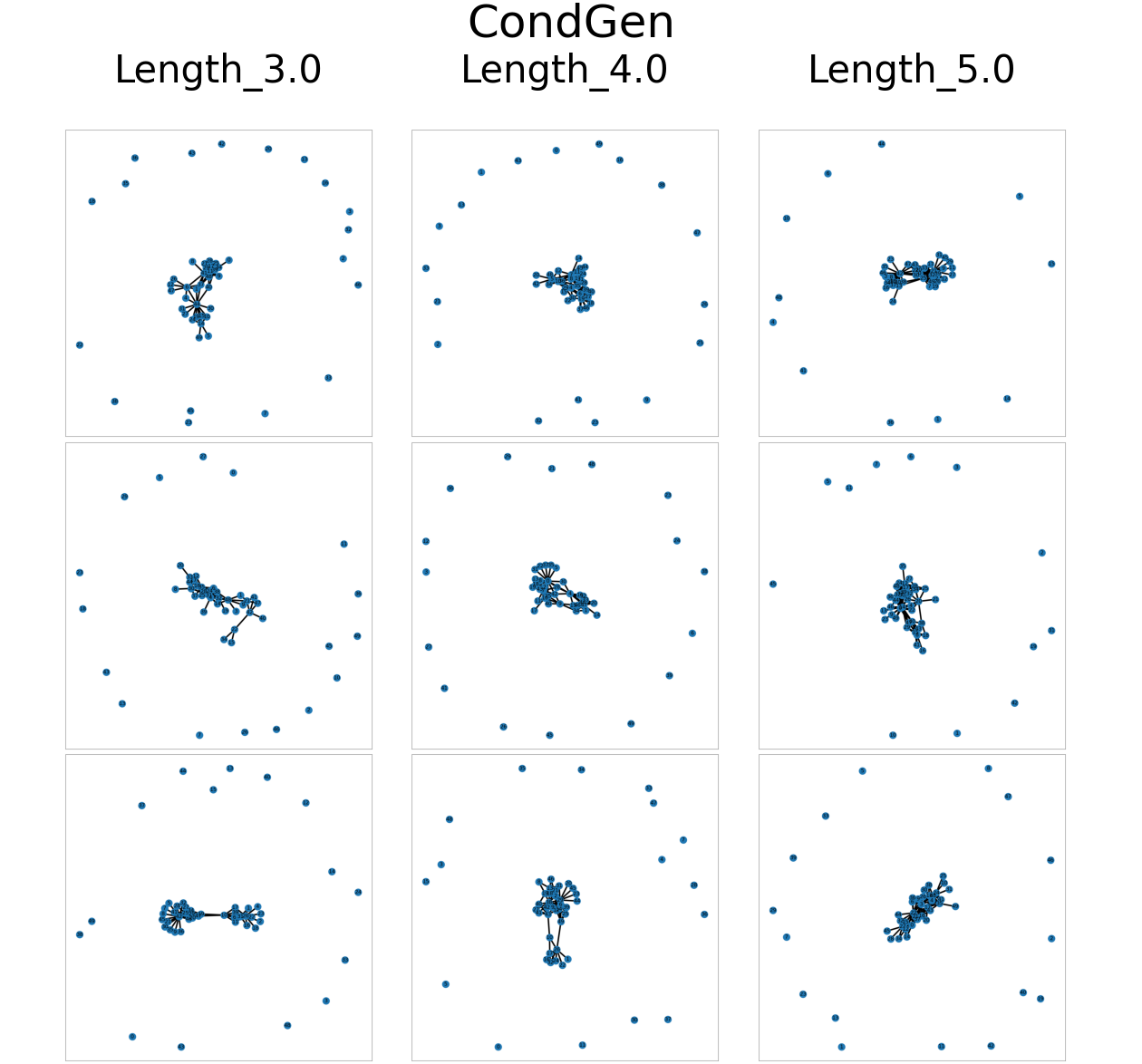}%
    \includegraphics[width=0.1\linewidth]{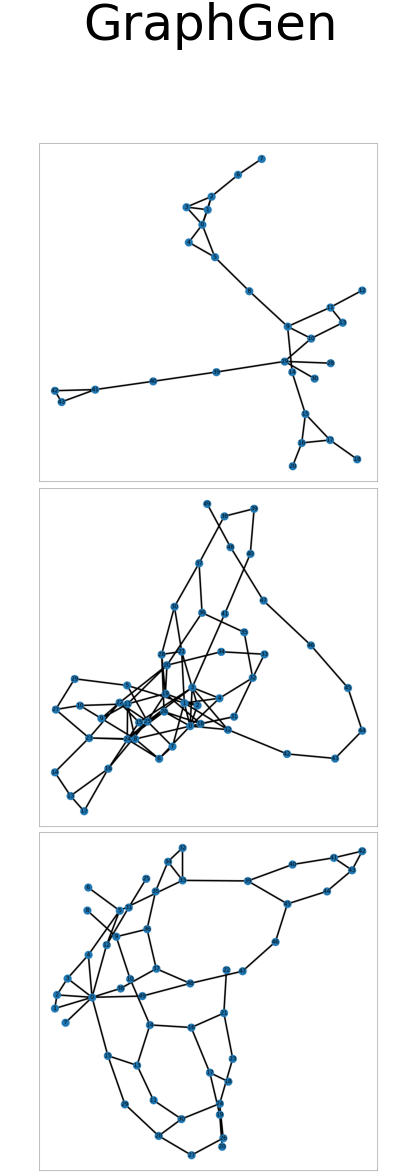}%
    % \vspace{-3ex}
  \end{center}
  \caption{
    Real data on the Twitter dataset and generated graphs by each generative model. 
    For all generative models except GraphGen, three samples of generative graphs when we specify the shortest path length to 3, 4, and 5 are depicted. 
    For real data, three samples whose average shortest path lengths are close to 3.0, 4.0, and 5.0 in the dataset are depicted. 
    In Real data and GraphTune, the graphs with ``length\_3'' clearly have a hub node, while the graphs with ``Length\_5'' forms a long loop.
    % ■■■ キャプションを書く ■■■  済
    % ■■■ GraphGenの結果を差し替え ■■■  済
  }
  \label{fig:sample}
\end{figure*}

According to the results in TABLE~\ref{tab:average}, we can confirm that the graphs generated by GraphTune have high reproduction accuracy and clearly change depending on the conditions. 
GraphGen generally reproduces real data well, but GraphTune, which adopts sequence-based generation like GraphTune, has similar performance.
A similar conclusion can be drawn from the visualized graphs of each model shown in Fig.~\ref{fig:sample}.
In the results of average degree and clustering coefficient, information of condition vector works effectively, and GraphTune has better reproduction accuracy than GraphGen.
In tunability of feature, GraphTune achieves the best performance in most of the features.
While GraphTune can accurately tune the average of shortest path length, the value for CondGen cannot be calculated since all generated graphs are unconnected.
Since we explicitly give information on the number of nodes and edges in the graph of the training dataset, CondGen can accurately tune the value of average degree.
GraphTune is also quite accurate even though such information is not given.
With regard to modularity and clustering coefficients, GraphTune outperforms CondGen in terms of both reproduction accuracy of real data and tunability.
For highly complex statistics such as the power-law exponent, the value of the feature is somewhat tunable but the result is a little unstable.
% ■■■ 可視化データの参照 ■■■  済

% Unfortunately, for highly complex statistics such as the Power-law exponent, it is difficult to tune. 
% However, we succeed to generate graphs with a similar feature that frequently appears in Real data.
% 追従できていない特徴量に関する言い訳や，特にうまく追従している特徴量の理由を書く．
% CondGenは，事前にラベルでグループ分けしたデータセットで学習する必要があるため，ラベルの一つをトレーニング後に柔軟にconditionの値をチューニングすることはできない．と書く．
% 結果の表で，CondGenが--となっている説明を書く．

\begin{table*}[tb]
  \caption{
    Average values of 5 global-level structural features in graphs generated by GraphTune, CondGen, and GraphGen for Twitter dataset. 
    We present the same value for all results of GraphGen for each feature since GraphGen does not provide conditional generation of graphs.
    The best performance between GraphTune and CondGen achieved under each condition for a particular feature is highlighted in bold font. 
    }
  \label{tab:average}
  \centering
   \begin{tabular}{rc|c|c|c|c|}
    Global-level structural feature & Condition & GraphTune & CondGen & GraphGen & Real data \\
    & & average & average & average & average \\
    & & & & & (25-percentile / median / 75-percentile) \\\hline\hline
    & 3.0   & \bf 3.05   &     -- ( 2.18 ) &        &       \\\cline{2-4}Average of shortest path length 
    & 4.0   & \bf 4.02   &     -- ( 2.24 ) & 4.59   & 4.26  \\\cline{2-4}
    & 5.0   & \bf 5.43   &     -- ( 2.27 ) &        & (3.40 / 4.09 / 4.84)      \\\hline
    & 3.0   &     3.26   & \bf 2.93   &        &       \\\cline{2-4}Average degree 
    & 3.5   &     3.60   & \bf 3.48   & 2.96   & 3.59  \\\cline{2-4}
    & 4.0   & \bf 3.90   &     4.51   &        & (2.96 / 3.44 / 3.92)    \\\hline
    & 0.40  & \bf 0.389  &     0.299  &        &       \\\cline{2-4}Modularity 
    & 0.55  & \bf 0.430  &     0.325  & 0.567  & 0.550 \\\cline{2-4}
    & 0.70  & \bf 0.507  &     0.336  &        & (0.509 / 0.563 / 0.617)      \\\hline
    & 0.1   & \bf 0.177  &     0.344  &        &       \\\cline{2-4}Clustering coefficient 
    & 0.2   & \bf 0.181  &     0.366  & 0.0846 & 0.203 \\\cline{2-4}
    & 0.3   & \bf 0.217  &     0.409  &        & (0.152 / 0.196 / 0.251)      \\\hline
    & 2.6   & \bf 2.91   &     4.10   &        &       \\\cline{2-4}Power-law exponent of a degree distribution
    & 3.0   & \bf 2.98   &     3.90   & 5.48   & 4.28  \\\cline{2-4}
    & 3.4   & \bf 3.50   &     3.80   &        & (2.91 / 3.48 / 4.23)      \\\hline
    % & 0.050 & \bf        &     0.0573 &        &       \\\cline{2-4}Edge density 
    % & 0.075 & \bf        &     0.0731 & 0.0623 & 0.0732\\\cline{2-4}
    % & 0.10  & \bf        &     0.105  &        &       \\\hline
    % & 10    &            &     --     &        &       \\\cline{2-4}Diameter
    % & 20    &            &     --     & 12.647 & 11.4  \\\cline{2-4}
    % & 30    &            &     --     &        &       \\\hline
   \end{tabular}
\end{table*}

\begin{figure*}[tbp]
  \begin{center}
    \includegraphics[width=0.8\linewidth]{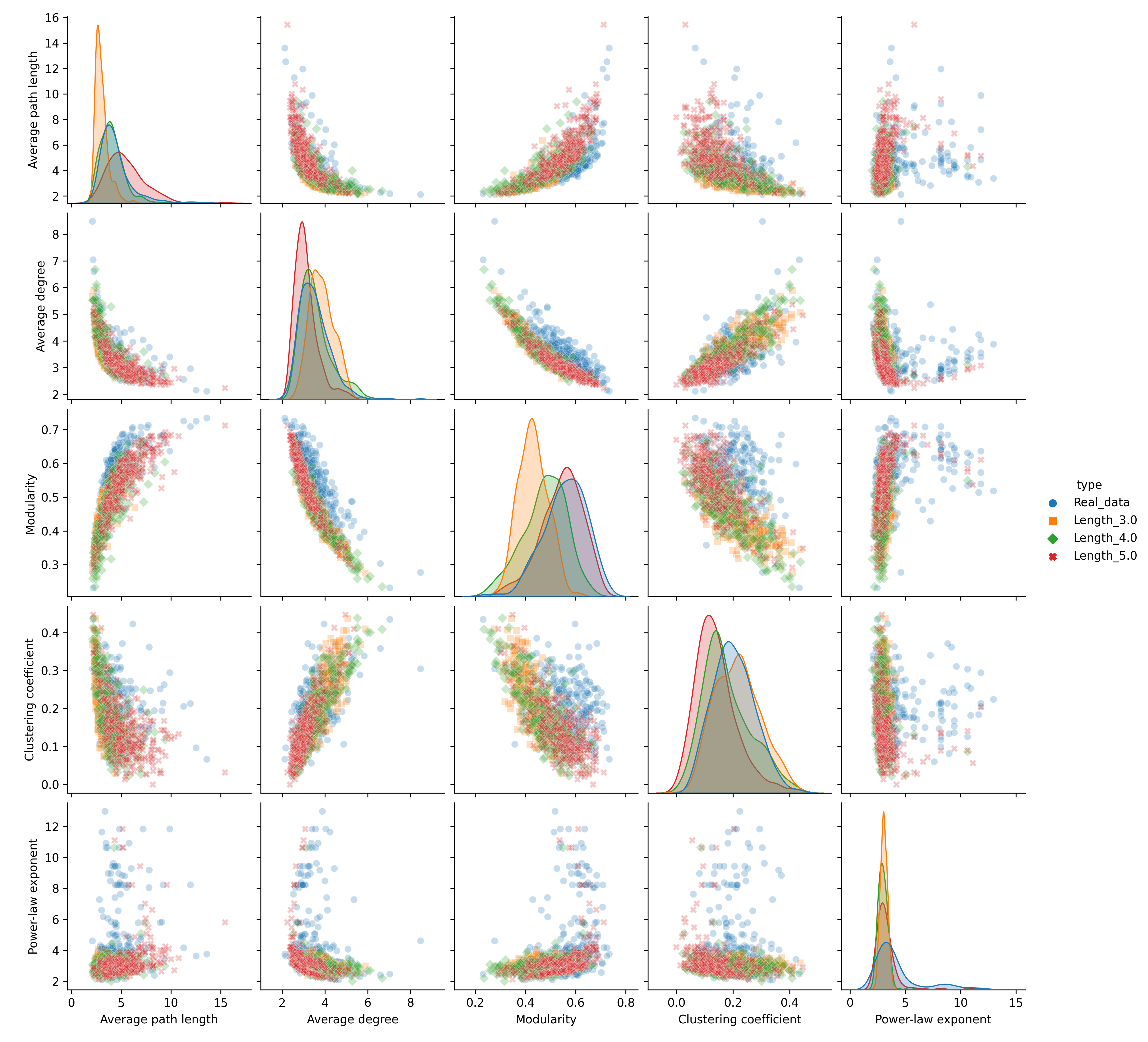}
  \end{center}
  \caption{
    Pairwise relationships of the values of the features on generated graphs by GraphTune when we give values 3.0, 4.0, and 5.0 of an average shortest path length as conditions. 
    The figure is a grid of multiple plots, and the grid is such that each feature will be shared across the y-axes across a single row and the x-axes across a single column. 
    The diagonal plots are the distributions of the features, and the others are scatter plots of two features.
    The distributions of values of an average shortest path length are clearly distinguishable.
  }
  % はっきりピークが分かれていること，複数の統計量の関係を忠実に再現していること，独立性の低い統計量同士は，調整対象の統計量に引きずられて変化していることを書く．
  \label{fig:pairplot}
\end{figure*}

To investigate the detailed performance of GraphTune, we depict the distributions of the values of the global-level structural features.
In Fig.~\ref{fig:pairplot}, we plot pairwise relationships of the values of the features on generated graphs by GraphTune. 
While maintaining the value of the other features of the generated graphs are in the range of real data, the distributions of values of an average shortest path length on GraphTune results are clearly distinguishable.
According to the scatter plots in Fig.~\ref{fig:pairplot}, we can confirm that the distribution of points in the real graph data and that of the generated graph data almost overlap. 
As a result, the relationships between any two features are accurately reproduced, and it was achieved that the reproduction of a graph dataset in every single aspect. 
 % 詳しい性能をinvestigateするために，分布を描いて分析する．
 % 分布がきちんと分化していて，ピークの位置が変わっている．ただ，強い依存関係にあるものは
 % ある程度独立に決められる特徴量同士なら，互いの依存関係が再現されていることをかく．特定のパラメータのみを制御することは難しくないが，互いの依存関係を再現できるのは，構造的特徴を多面的に再現していることの証左．

\subsection{Ablation Study}
\label{ssec:AblationExperiment}

We performed an ablation study to demonstrate the validity of the architecture of GraphTune.
When giving a condition to the LSTM-based VAE in GraphTune, a condition vector is input to 3 spots: the input sequence of the encoder (Eq.~(\ref{eq:enc_cond})), the input sequence of the decoder (Eq.~(\ref{eq:dec_cond})), and the hidden layer of the decoder (Eq.~(\ref{eq:hidden_cond})).
To evaluate the impact of the condition vector on the tunability, we prepared 3 versions of GraphTune by removing one of the 3 spots at which a condition vector is input.
We trained each model with Twitter dataset and generated 300 graphs.
Each model is trained with the same parameters as the experiment regarding the average shortest path length shown in Section~\ref{ssec:PerformanceEvaluations}. 
We calculated the average shortest path length for all generated graphs, and calculated the Root Mean Squared Error (RMSE) with the value specified in the condition vector as the true value.
The results of RMSE for each model are listed in TABLE~\ref{tab:ablation}.
As shown in TABLE~\ref{tab:ablation}, RMSE of the original GraphTune is the lowest value among all models.
According to the results, although 3 spots of input of a condition vector in the GraphTune architecture seem redundant, we can understand that all condition inputs contribute to improving its tunability.
The most significant contributor to the tunability is the condition vector that is concatenated to the input sequence of the encoder.
This suggests that removing the information related to the specified feature from the latent vector is important for the graph generation.
Meanwhile, the condition vector that is concatenated to the input sequence of the decoder also contributes significantly to the tunability.
% ■■■ 追加するか検討 ■■■ This is consistent with the result that the GraphGen with a condition vector evaluated in Section~\ref{ssec:PerformanceEvaluations} has somewhat tunability.
% ■■■ 切除テスト ■■■  済

\begin{table*}[tb]
  \caption{
  RSME for each GraphTune version. 
  We verify 3 versions by removing one of the 3 spots at which a condition vector is input. 
  The smallest values of RMSE among all models achieved under each condition are highlighted in bold font.
    % The average shortest path length in graphs generated by GraphTune, CondGen, and GraphGen for WS dataset. 
    % We present the same value for all results of GraphGen for each feature since GraphGen does not provide conditional generation of graphs.
    % The best performance between GraphTune and CondGen achieved under each condition for a particular feature is highlighted in bold font. 
    }  
  \label{tab:ablation}
  \centering
   \begin{tabular}{rc|c|c|c|c|}
    Global-level structural feature & Condition & GraphTune & GraphTune without & GraphTune without & GraphTune without \\
    & & Original & condition on encoder & condition on decoder & condition on hidden layer \\
    & & RMSE & RMSE & RMSE & RMSE \\\hline\hline
    & 3.0 & \bf 0.70 & 3.06 & 2.63 & 1.12 \\\cline{2-6} Average of shortest path length 
    & 4.0 & \bf 1.15 & 3.22 & 2.82 & 1.38 \\\cline{2-6}
    & 5.0 & \bf 1.92 & 4.17 & 3.47 & \bf 1.92 \\\hline
    % &    &       &   &  &       \\\cline{2-4}Average degree 
    % &    &       &   &  &   \\\cline{2-4}
    % &    & \bf   &   &  & ( /  / )    \\\hline
    % &    & \bf   &   &  &       \\\cline{2-4}Modularity 
    % &    & \bf   &   &  &  \\\cline{2-4}
    % &    & \bf   &   &  & ( /  / )      \\\hline
    % &    & \bf   &   &  &       \\\cline{2-4}Clustering coefficient 
    % &    & \bf   &   &  &  \\\cline{2-4}
    % &    & \bf   &   &  & ( /  / )      \\\hline
    % &    & \bf   &   &  &       \\\cline{2-4}Power-law exponent of a degree distribution
    % &    & \bf   &   &  &   \\\cline{2-4}
    % &    & \bf   &   &  & ( /  / )      \\\hline
    % & 0.050 & \bf        &     0.0573 &        &       \\\cline{2-4}Edge density 
    % & 0.075 & \bf        &     0.0731 & 0.0623 & 0.0732\\\cline{2-4}
    % & 0.10  & \bf        &     0.105  &        &       \\\hline
    % & 10    &            &     --     &        &       \\\cline{2-4}Diameter
    % & 20    &            &     --     & 12.647 & 11.4  \\\cline{2-4}
    % & 30    &            &     --     &        &       \\\hline
   \end{tabular}
\end{table*}

\subsection{Latent Space Analysis}

Together with the above evaluations regarding tunability, we analyzed the structure of the latent state distribution $\mathcal{Z}$ in GraphTune.
The reference~\cite{Stoehr2019} focuses to enforce disentangled representations of model parameters in a graph generation based on the idea of $\beta$-VAE~\cite{Higgins2017}.
Although the proposed model in the reference~\cite{Stoehr2019} cannot provide conditional generation of graphs, the authours reported that it is possible to construct a generative model that correlates the number of nodes $n$ and edge density $p$ with the elements of the latent vector $\boldmath{z}$ in a simple ER model. 
We performed an experiment with the graph dataset (ER dataset) generated by the same conditions as ER model in reference~\cite{Stoehr2019}, and verified the structure of the latent space and generation results.
All parameters in GraphTune are the same as those evaluated in Section~\ref{ssec:PerformanceEvaluations}, except the dimension of the latent space is set to 4 which is the same as the experimental conditions in reference~\cite{Stoehr2019}.
In the learning process, we used connected graphs in ER dataset, and gave edge density $p$ as a condition vector.

\begin{figure}[tb]
  \begin{center}
    \includegraphics[width=\linewidth]{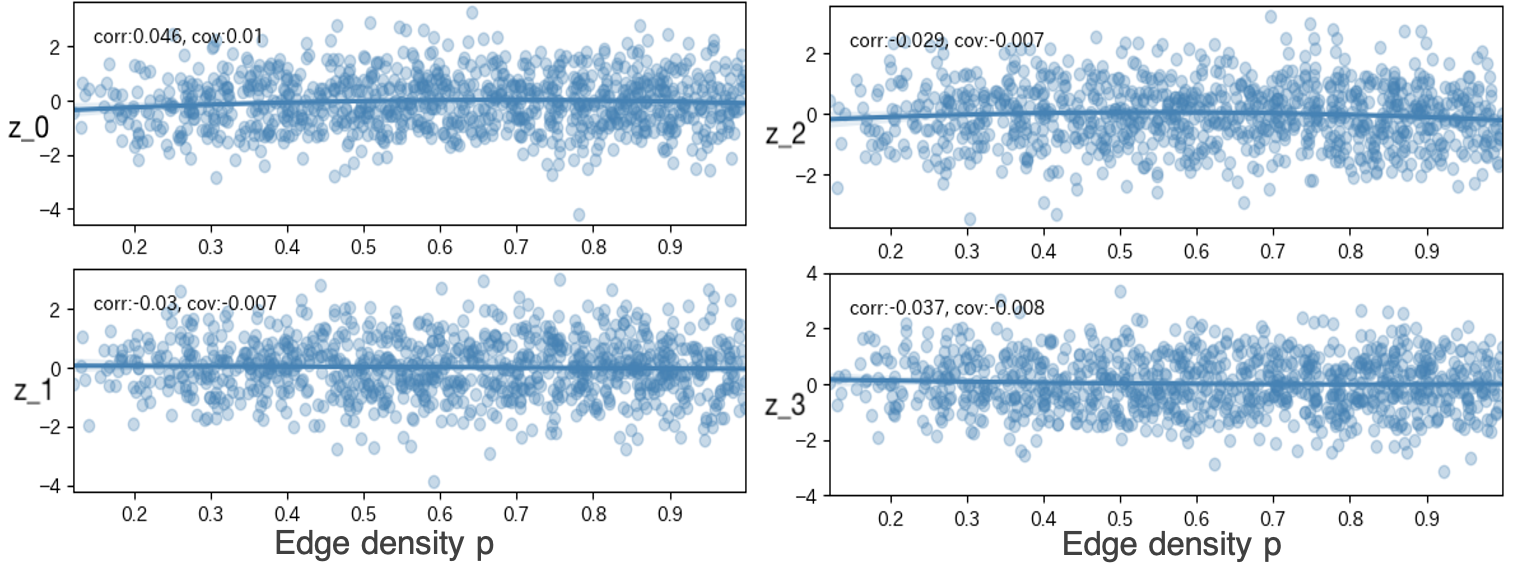}
  \end{center}
  \caption{
    Relationship between elements $z_i$ of a latent vector $\boldsymbol{z}$ and edge density $p$ of ER model.
    It is confirmed that the latent space of GraphTune is not disentangled.
    % ■■■ キャプション書き換え ■■■
  }
  % はっきりピークが分かれていること，複数の統計量の関係を忠実に再現していること，独立性の低い統計量同士は，調整対象の統計量に引きずられて変化していることを書く．
  \label{fig:er_latent_space}
\end{figure}

\begin{figure}[tb]
  \begin{center}
    \includegraphics[width=0.8\linewidth]{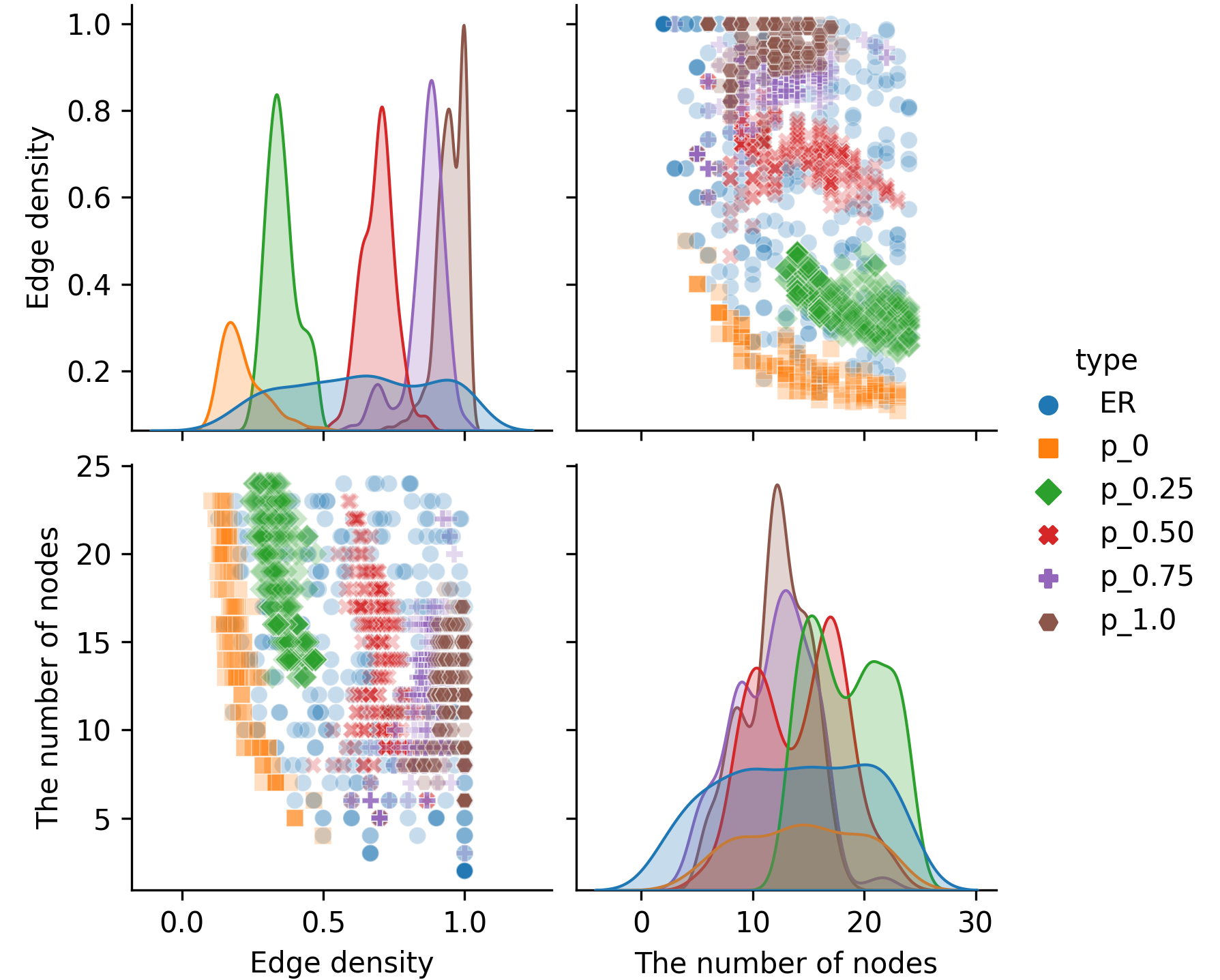}
  \end{center}
  \caption{
    Pairwise relationships of the values of the features on generated graphs by GraphTune when we give values 0.0, 0.25, 0.5, 0.75, and 1.0 of an edge density $p$ as conditions. 
    The distributions of $p$ are clearly changed depending on a value of a condition vector while the distribution is not significantly dependent on a value of the number $n$ of nodes. 
    % ■■■ キャプション書き換え ■■■  済
  }
  % はっきりピークが分かれていること，複数の統計量の関係を忠実に再現していること，独立性の低い統計量同士は，調整対象の統計量に引きずられて変化していることを書く．
  \label{fig:er_pairplot}
\end{figure}

Fig.~\ref{fig:er_latent_space} shows the relationship between each element $z_i$ of the latent vector $\boldsymbol{z}$ and edge density $p$, and it is confirmed that the latent space of GraphTune is not disentangled according to the figure. 
This analysis that visualizes the correlation between $\boldsymbol{z}$ and $p$ is adopted in the evaluation in reference~\cite{Stoehr2019}.
On the other hand, from the pairwise relationship of $n$ and $p$ shown in Fig.~\ref{fig:er_pairplot}, we can clearly confirm the correlation between values of condition vectors and edge density $p$ of the generated graphs. 
In this result, the correlation coefficient between values of the condition vector and $p$ is 0.95, and it is higher than the correlation coefficient of 0.35 between the element of latent vector $z_i$ and $p$ reported in the reference~\cite{Stoehr2019}.
It should be noted that $p$ depends on a condition, whereas $n$ can be determined independently of a condition. 
The above results mean that GraphTune achieves to generate graphs depending only on the value of the condition given to the decoder, not on the information of the latent space. 

% ■■■ Disentangled vector の検証 ■■■  済
% ■■■ \cite{xxx}が調整不能モデルであることをrelated workに書く ■■■  済

\section{Limitations and Future Directions}\label{sec:Limitations}

We recognize that there are some limitations of generation by GraphTune, which suggest that GraphTune has potential for future expansion. 

\subsubsection*{Generation of Large Graphs}

Although the number of nodes on graphs in our dataset is relatively large compared with the datasets that have been evaluated in studies regarding learning-based conditional generations of graphs, it is small in terms of social networks. 
In the results~\cite{Goyal2020} of unconditional generation with GraphGen, which adopts sequence-based generation like GraphTune, the average number of nodes of graphs generated by GraphGen is at most 54.01 nodes. 
While GraphTune generates graphs of almost the same size as the graphs that GraphGen generates, it was unfortunately not able to generate graphs with over 100 nodes. 
More innovation is needed to overcome this limitation. 
Hierarchical generation~\cite{Jin2020} is relatively easy to implement but is expected to be effective and is a promising option.
Another promising option is a combination of deep learning and traditional statistical graph generation. 
We consider that the sequence-based generation in GraphTune has a high affinity and extensibility for both approaches. 

%我々は，比較的小さなサイズのネットワークのGlobal-level featuresの調整に成功したに過ぎません．

\subsubsection*{Pinpoint Specification of Features}

The tunability of graph features in GraphTune is not perfect. 
While a distribution of feature values of graphs generated by GraphTune has distinctly different peaks, these values are somewhat varied. 
Accuracy improvements of the specification remain as a future issue. 
In addition, GraphTune can specify at most one feature, and currently, it has not succeeded in specifying multiple features at the same time.
In order to achieve the specification of multiple features, it is necessary to understand the independency and/or dependency between each feature.
However, the independency and/or dependency of global-level structural features is very complex, and understanding it is a challenging issue.
It goes without saying that analytical results based on graph theory are important for this issue. 
However, observing the features of graphs generated by GraphTune may reinforce the results of graph theory by a data-driven approach. 

% 依存関係にあるもの別の特徴量を完全に固定することはできない．そもそも依存関係は明確ではないので，それを明らかにすることがこれらの問題解決のよいアプローチになると思われる．

\subsubsection*{Generation of Extrapolation}

Although GraphTune generates graphs with a specific feature flexibly, the tunable range of a feature is limited to the range of the feature within the graphs of the learned dataset.
In the tasks of generation or prediction, it is generally hard to extrapolate values that are not included in the range of the dataset.
This difficulty is the same for the graph generation task, and the current GraphTune cannot generate extrapolated graphs that are outside the range of graphs included in the training set. 
However, the specification technique of graph features provided by GraphTune could be a key technology to overcome the difficulty of extrapolation output in the graph generation task. 
By specifying the value of a feature on the edge of the range of the training dataset, we can generate graphs inside and outside the border.
The generated graphs enhance the original training set, and the enhanced training set covers ranges that are not included in the original training set by repeatedly generating graphs on the edges.
The generation of extrapolated graphs is one of our future directions.

% 外挿はできないとかそういう話をする．
% 特徴量を訓練データセットのレンジのエッジに指定することで，境界線上の内側と外側のデータを生成することができます．

\section{Conclusion}\label{sec:Conclusion}

In this work, we proposed GraphTune, which is a learning-based graph generative model with tunable structural features. 
GraphTune is composed of a CVAE with an LSTM-based encoder and decoder. 
By specifying the value of a particular structural feature as a condition vector that is input into CVAE, we can generate graphs with a specific structural feature. 
We performed comparative evaluations of GraphTune, CondGen, and GraphGen through a real graph dataset sourced from the who-follows-whom graph on Twitter. 
The result of the evaluation show that GraphTune makes it possible to tune the value of a global-level structural feature, and that conventional models are unable to tune global-level structural features.

Although GraphTune provides a rich variety of graphs flexibly, it does not solve all problems related to graph modeling.
One improvement needed in future works is to provide rich functionality for the specification of structural features. 
In addition to improving the accuracy of feature values of generated graphs, it is also necessary to be able to specify multiple features at the same time. 
Allowing the generation of extrapolated graphs that are not included in the training dataset is also an important function.
On the other hand, the tunability of GraphTune has the potential to empower traditional graph theory by a data-driven approach.
When combined with traditional graph theory, unraveling complex global-level features relationships is a challenging but interesting issue.

% 外装の予測に使えるという話を書く．

% or


% \appendices
% % you can choose not to have a title for an appendix
% % if you want by leaving the argument blank
% \section{}

% % use section* for acknowledgment
% \ifCLASSOPTIONcompsoc
%   % The Computer Society usually uses the plural form
%   \section*{Acknowledgments}
% \else
%   % regular IEEE prefers the singular form
% \fi
% 
% 
% This work was partly supported by JSPS KAKENHI Grant Number JP20H04172.  

\ifCLASSOPTIONcaptionsoff
  \newpage
\fi



\bibliographystyle{IEEEtran}
\bibliography{IEEEabrv,list,book_bib}
\end{document}